\documentclass[preprint,5p,compress]{elsarticle}

\usepackage{lscape,caption}
\usepackage{amssymb}
\usepackage{amsmath}
\usepackage{mathtools}
\usepackage{graphicx, subfigure}
\usepackage{url}
\usepackage{algorithm}
\usepackage{algpseudocode}
\usepackage{color}
\usepackage{booktabs}
\usepackage{stmaryrd}
\usepackage{epstopdf}
\usepackage{color}

\journal{Neurocomputing}

\definecolor{mycolor}{rgb}{0,0,0} % Chane to (0,0,0) to remove highlighting of updated text

\begin{document}

\begin{frontmatter}

\title{Dealing with Difficult Minority Labels in Imbalanced Mutilabel Data Sets}

\author[UGR]{Francisco Charte\corref{cor1}}
\ead{francisco@fcharte.com}

\author[UJA]{Antonio J. Rivera}
\ead{arivera@ujaen.es}

\author[UJA]{Mar\'ia J. del Jesus}
\ead{mjjesus@ujaen.es}

\author[UGR]{Francisco Herrera}
\ead{herrera@ugr.es}

\cortext[cor1]{Corresponding author. Tel.: +34 953 212 892; fax: +34 953 212 472. \\ %Arxiv
	Manuscript accepted at Neurocomputing: \url{https://doi.org/10.1016/j.neucom.2016.08.158} \\
	\textcopyright~2018. This manuscript version is made available under CC  	BY-NC-ND 4.0 license \url{https://creativecommons.org/licenses/by-nc-nd/4.0/}}
\address[UGR]{Department of Computer Science and A.I., University of Granada, 18071 Granada, Spain}
\address[UJA]{Department of Computer Science, University of Ja\'en, 23071 Ja\'en, Spain}

\begin{abstract}
Multilabel classification is an emergent data mining task with a broad range of real world applications. Learning from imbalanced multilabel data is being deeply studied latterly, and several resampling methods have been proposed in the literature. The unequal label distribution in most multilabel datasets, with disparate imbalance levels, could be a handicap while learning new classifiers. In addition, this characteristic challenges many of the existent preprocessing algorithms. Furthermore, the concurrence between imbalanced labels can make harder the learning from certain labels. These are what we call \textit{difficult} labels. In this work, the problem of difficult labels is deeply analyzed, its influence in multilabel classifiers is studied, and a novel way to solve this problem is proposed. Specific metrics to assess this trait in multilabel datasets, called \textit{SCUMBLE} (\textit{Score of ConcUrrence among iMBalanced LabEls}) and \textit{SCUMBLELbl}, are presented along with REMEDIAL (\textit{REsampling MultilabEl datasets by Decoupling highly ImbAlanced Labels}), a new algorithm aimed to relax label concurrence. How to deal with this problem using the R mldr package is also outlined.
\end{abstract}


\end{frontmatter}

\section{Introduction}
Multilabel classification (MLC) \cite{Charte:SB-MLC,TutorialVentura} models are designed to predict the subset of labels associated to each instance in a multilabel dataset (MLD), instead of only one class as traditional classifiers do. It is a task useful in fields such as automated tag suggestion \cite{QUINTA}, protein classification \cite{genbase}, and object recognition in images \cite{Wei2013}, among others. Many different methods have been proposed lately to accomplish this problem.

The number of instances in which each label appears is not homogeneous. In fact, most MLDs show big differences in label frequencies. This peculiarity is known as imbalance \cite{Chawla:2004}, and it has been profoundly studied in traditional classification. In the context of MLC, several proposals to deal with imbalanced MLDs \cite{Tepvorachai:2008,He:2012,Tahir:2012:2,Tahir:2012,LISHI:2013,Charte:Neucom13,Giraldo:2013,Charte:IDEAL14,Charte:MLSMOTE} have been made lately. Despite these efforts, there are still some aspects regarding imbalanced learning in MLC that would need additional analysis.

Resampling techniques are commonly used in with traditional (non-multilabel) datasets \cite{JSSanchez:13} to balance their class distributions, hence they are an obvious choice to face the same problem with MLDs. Notwithstanding, the nature of MLDs can be a challenge for resampling algorithms. In this paper we will show how a specific characteristic of these datasets, the joint presence in the same instance of labels with different frequencies, could prevent the goal of these algorithms. 

We hypothesized\footnote{This paper is an extended version of our previous work \cite{Charte:HAIS14} from HAIS'14, including additional metrics, a deeper analysis, and an algorithm aimed to solve the described problem. The proposed solutions have been implemented in R, and the software package containing them is also described.} that this symptom, the concurrence among imbalanced labels, would influence the resampling algorithms behavior. Specifically, the minority labels which jointly appear with majority ones would be difficult labels. In order to deal with this problem we propose to face it in two phases:
\begin{itemize}
    \item Firstly, the concurrence problem has to be assessed. For doing so, two new metrics, named \textit{SCUMBLE} (\textit{Score of ConcUrrence among iMBalanced LabEls}) and \textit{SCUMBLELbl}, designed explicitly to assess this causality, will be proposed. Its effectiveness will be experimentally demonstrated.
    
    \item Secondly, an algorithm specifically designed to preprocess MLDs affected by this problem would be needed. A such method, called REMEDIAL (\textit{REsampling MultilabEl datasets by Decoupling highly ImbAlanced Labels}), will be introduced, and its performance will be empirically tested.
\end{itemize} 

The \textit{SCUMBLE} measure was conceived aiming to know how difficult would be to work with a certain MLD for resampling algorithms. Its goal is to appraise the concurrence among imbalanced labels, giving as result a score easily interpretable. This score will be in the range [0,1]. A low score would denote an MLD with not much concurrence among imbalanced labels, whereas a high one would evidence the opposite case. Our hypothesis is that the lower the score obtained, the better the resampling algorithms would work. \textit{SCUMBLELbl} complements the former metric, allowing to know which labels are more affected by this problem. The less frequent labels with a high \textit{SCUMBLELbl} would be specially difficult cases.

Once the presence of the concurrence problem has been stated, the idea of how to deal with it naturally arises. For this reason the algorithm REMEDIAL, a specific method able to reduce concurrence among imbalanced labels, is also introduced. REMEDIAL works by decoupling imbalanced labels through an editing and oversampling approach. It is a resampling algorithm, since it produces new data samples. At the same time, it also edits existent instances. \textcolor{mycolor}{However, it does not change the number of times that each label appears in the dataset.} The details about REMEDIAL and how to use this algorithm along with the proposed metrics, relying on the mldr R package \cite{Charte:mldr}, will also be explained.

The rest of this paper is structured as follows. Section \ref{Preliminaries} offers a brief introduction to MLC, as well as a description on how the learning from imbalanced MLDs has been faced. In Section \ref{ImbalanceMLC} the problem of concurrence among imbalanced labels in MLDs will be defined, and how to assess this concurrence using the proposed metrics will be explained. The algorithm REMEDIAL is described in Section \ref{REMEDIAL}. Section \ref{Experimentation} portraits the experimental framework used, as well as the obtained results from experimentation. Finally, Section \ref{Conclusions} will offer the conclusions. \textcolor {mycolor}{Appendix \ref{mldrPackage} describes how to assess the label concurrence level and how to apply the REMEDIAL algorithm thorough a specific software package developed by the authors.}

\section{Preliminaries}\label{Preliminaries}
In this section a concise introduction to multilabel classification is offered, along with a description on how the learning from imbalanced MLDs has been faced until now.

\subsection{Multilabel Classification}
Currently, there are many domains \cite{genbase,Wei2013,CAL500,enron,Elisseeff1,medical} in which each \textcolor {mycolor}{data pattern} is not associated \textcolor {mycolor}{exclusively to one} class, but to a group of them. In this context the classes are named labels, and the set of labels that belongs to a data sample is called labelset. Let $D$ be an MLD, $D_i$ the \textit{i-th} instance, and $L$ the full set on labels on $D$. The goal of a multilabel classifier is to predict a set $Z_i \subseteq L$ with the labelset for $D_i$.

Multilabel classification has been traditionally faced through two different approaches \cite{Tsoumakas3}. The first one, called data transformation, aims to produce binary or multiclass datasets from an  MLD, allowing the use of non-MLC algorithms. The second, known as algorithm adaptation, has the goal of adapting established algorithms to natively work with MLDs. The two most common transformation methods are Binary Relevance (BR) \cite{Godbole} and Label Powerset (LP) \cite{Boutell}. The former produces several binary datasets from an MLD, usually one for each label or one for each label pair \cite{mencia2010efficient}. The latter transforms the MLD \textcolor {mycolor}{into} a multiclass dataset, taking each labelset as class identifier. Regarding adapted algorithms, the number of proposals is quite high. There are multilabel KNN classifiers such as ML-kNN \cite{Zhang1}, multilabel trees based on C4.5 \cite{Clare}, and multilabel ANNs such as \cite{Zhou:MIML:2009}, as well as a profusion of algorithms based on ensembles of BR and LP classifiers. A recent review on multilabel classification algorithms can be found in \cite{TutorialVentura}.

Thus far, most proposed multilabel characterization metrics are focused in assessing the number of labels and labelsets. The most common ones are the total number of labels $|L|$,  label cardinality ($Card$), which is the average number of labels per instance, and label density, obtained as $Card/|L|$.

\subsection{Learning from Imbalanced Data}\label{ImbalanceMetrics}
Imbalanced learning is a well-known problem in traditional classification \cite{Alberto:2013,He2009,Lopez:2013,Prati2014}, having been faced through three main approaches. First, by way of algorithmic adaptations of existent classifiers, the imbalance is taken into account in the classification process. Second, the preprocessing approach aims to balance class distributions by way of data resampling, creating (oversampling) or removing (undersampling) data samples. Third, cost sensitive classification is a combination of the two previous approaches. The data resampling approach has the advantage of being classifier independent, and its effectiveness has been proven in many scenarios.

In the MLC field, both the algorithmic adaptation and the data resampling approaches have been applied. The former is present in \cite{Tepvorachai:2008,He:2012,LISHI:2013}, while the latter appears in \cite{Tahir:2012,Charte:Neucom13,Giraldo:2013,Charte:IDEAL14,Charte:MLSMOTE}. There are also proposals based on the use of ensemble of classifiers, such as \cite{Tahir:2012:2}.

When it comes to assess\textcolor {mycolor}{ing} the imbalance level in MLDs, the metrics in Eq. \ref{IRLbl} and Eq. \ref{MeanIR} are proposed in \cite{Charte:HAIS13}. Let $D$ be an MLD, $L$ the full set of labels in it, $y$ the label being analyzed, and $Y_i$ the labelset of \textit{i-th} instance in $D$. In Eq. \ref{IRLbl} the symbol $\llbracket \rrbracket$ denotes de Iverson bracket, which returns 1 if the expression inside it is true or 0 otherwise. \textit{IRLbl} is a measure calculated individually for each label. The higher is the \textit{IRLbl} the larger would be the imbalance, allowing to know which labels are in minority or majority. \textit{MeanIR} is the average \textit{IRLbl} for an MLD. It is useful to estimate the global imbalance level.

\begin{equation} \textit{IRLbl(y)} = 
   \frac{
      \displaystyle\max\limits_{y' \in L}^{}
      	\left(\displaystyle\sum\limits_{i=1}^{|D|}{\llbracket y' \in Y_i \rrbracket}\right)
   }
   {
      \displaystyle\sum\limits_{i=1}^{|D|}{\llbracket y \in Y_i \rrbracket}} . 
   \label{IRLbl}\end{equation}

\begin{equation}
	\textit{MeanIR} = \frac{1}{|L|} \displaystyle\sum\limits_{y \in L}^{}\textit{IRLbl(y)} .
\label{MeanIR}\end{equation}
  
Even though the previously cited proposals for facing imbalanced learning in MLC achieve some good results, their behavior is heavily influenced by MLDs characteristics such as the imbalance levels, measured by means of the previous metrics, or the concurrence among \textcolor {mycolor}{imbalanced} labels, which will be described later. In the following we will focus in this topic, specifically in regard to data resampling solutions.

\subsection{Related Work}\label{ResamplingMethods}
In general, resampling methods aimed to work with non-MLDs can be divided into two categories, oversampling algorithms and undersampling algorithms. The former technique produces new samples with the minority class, while the latter removes instances linked to the majority class. The way in which the samples to be removed or reproduced are chosen can also be grouped into two categories, random methods and heuristic methods. Since this kind of datasets use only one class per instance, the previous techniques effectively balance the distribution of classes. However, this is not always true when dealing with MLDs. Moreover, most MLDs have more than one minority and one majority label.

The preceding approaches have been migrated to the multilabel scenario at some extent, giving as result proposals such as the following:
\begin{itemize}
	\item \textbf{Random undersampling}: Two multilabel random undersampling algorithms are presented in \cite{Charte:Neucom13}, one of them based on the LP transformation (LP-RUS) and another one on the \textit{IRLbl} measure (ML-RUS). The latter determines what labels are in minority, by means of their \textit{IRLbl}, and avoids removing samples in which they appear.
	
	\item \textbf{Random oversampling}: The same paper \cite{Charte:Neucom13} also proposes two random oversampling algorithms, called LP-ROS and ML-ROS. The former is based on the LP transformation, while the latter relies on the \textit{IRLbl} measure. Both take into account several minority labels, and generate new instances cloning the original labelsets.
	
	\item \textbf{Heuristic undersampling}: In \cite{Charte:IDEAL14} a method to undersample MLDs following the ENN (\textit{Edited Nearest Network}) rule was presented. The instances are not randomly chosen, as in LP-RUS or ML-RUS, but carefully selected after analyzing their \textit{IRLbl} and the differences with their neighborhood.
	
	\item \textbf{Heuristic oversampling}: The procedure proposed in \cite{Giraldo:2013} is based on the original SMOTE algorithm. First, instances of an MLD are chosen using different criteria, then the selected samples are given as input to SMOTE, producing new samples with the same labelsets. In \cite{Charte:MLSMOTE} a more sophisticated approach is presented, with a multilabel version of SMOTE, called MLSMOTE, able to produce synthetic samples whose labelsets are generated from those of the nearest neighbors, instead of cloning them.
\end{itemize}

A major disadvantage in some of these algorithms is that they always work over full labelsets, cloning the set of labels in existent samples or completely removing them. Although this approach can benefit some MLDs, in other cases the result can be counterproductive depending on the MLD traits.

The aforementioned multilabel resampling algorithms will not have an easy work while dealing with MLDs which have a high \textit{SCUMBLE} level. Undersampling algorithms can produce a loss of essential information, as the samples selected for removal because majority labels appear in them can also contain minority labels. In the same way, oversampling algorithms limited to cloning the labelsets, such as the proposals in \cite{Charte:Neucom13,Giraldo:2013}, can be also increasing the presence of majority labels.

\section{Imbalanced MLDs and Resampling Algorithms Behavior}\label{ImbalanceMLC}
Most traditional resampling methods do their job by removing instances with the most frequent class, or creating new samples from instances associated to the least frequent one. Since each instance can belong to one class only, these actions would effectively balance the classes frequencies. However, this is not necessarily true when working with MLDs.

\begin{figure*}[t]
\begin{center}
  \includegraphics[width=.75\textwidth]{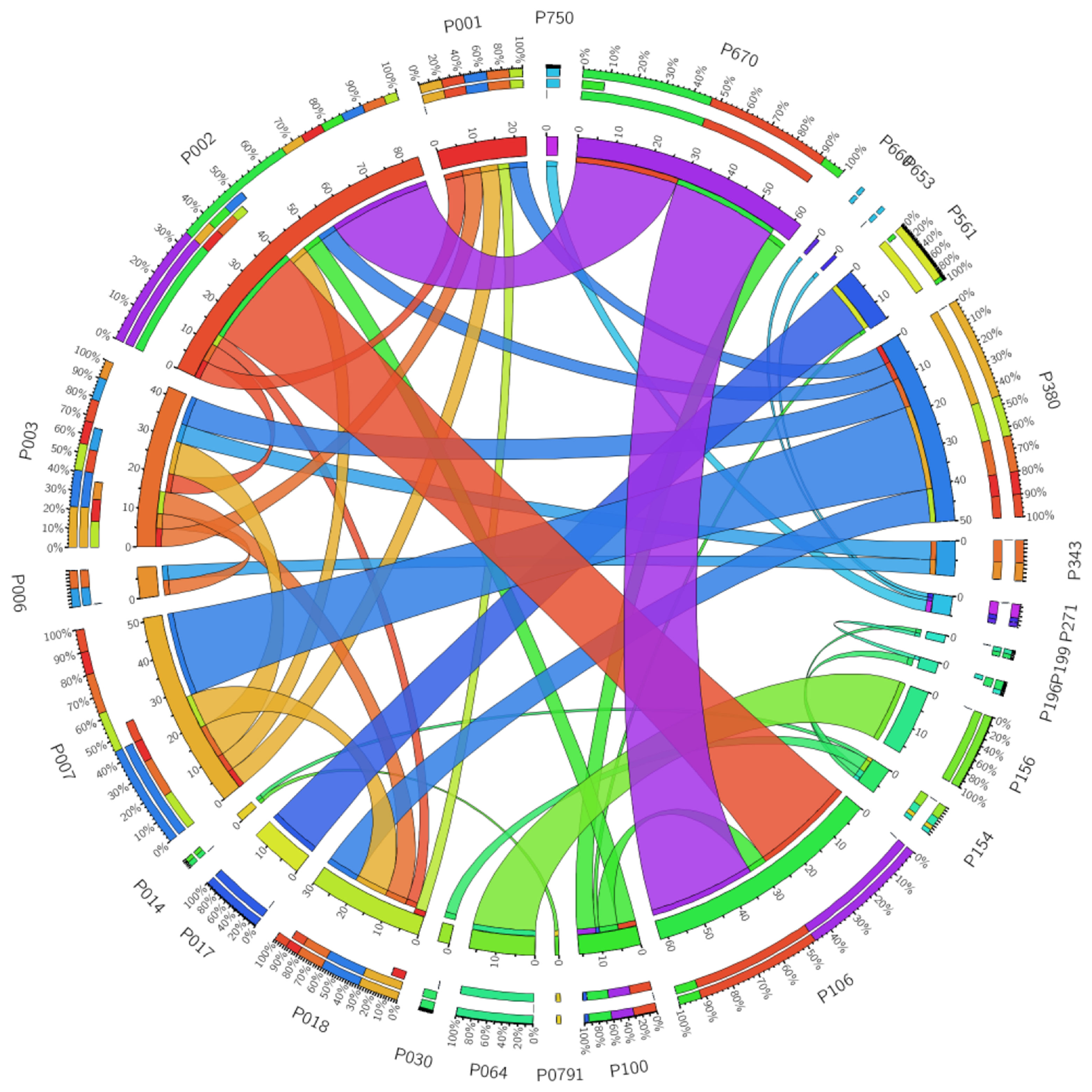}
\end{center}
\caption{Label concurrence in genbase MLD.} \label{genbase}
\end{figure*}

\begin{figure*}[t]
\begin{center}
  \includegraphics[width=.75\textwidth]{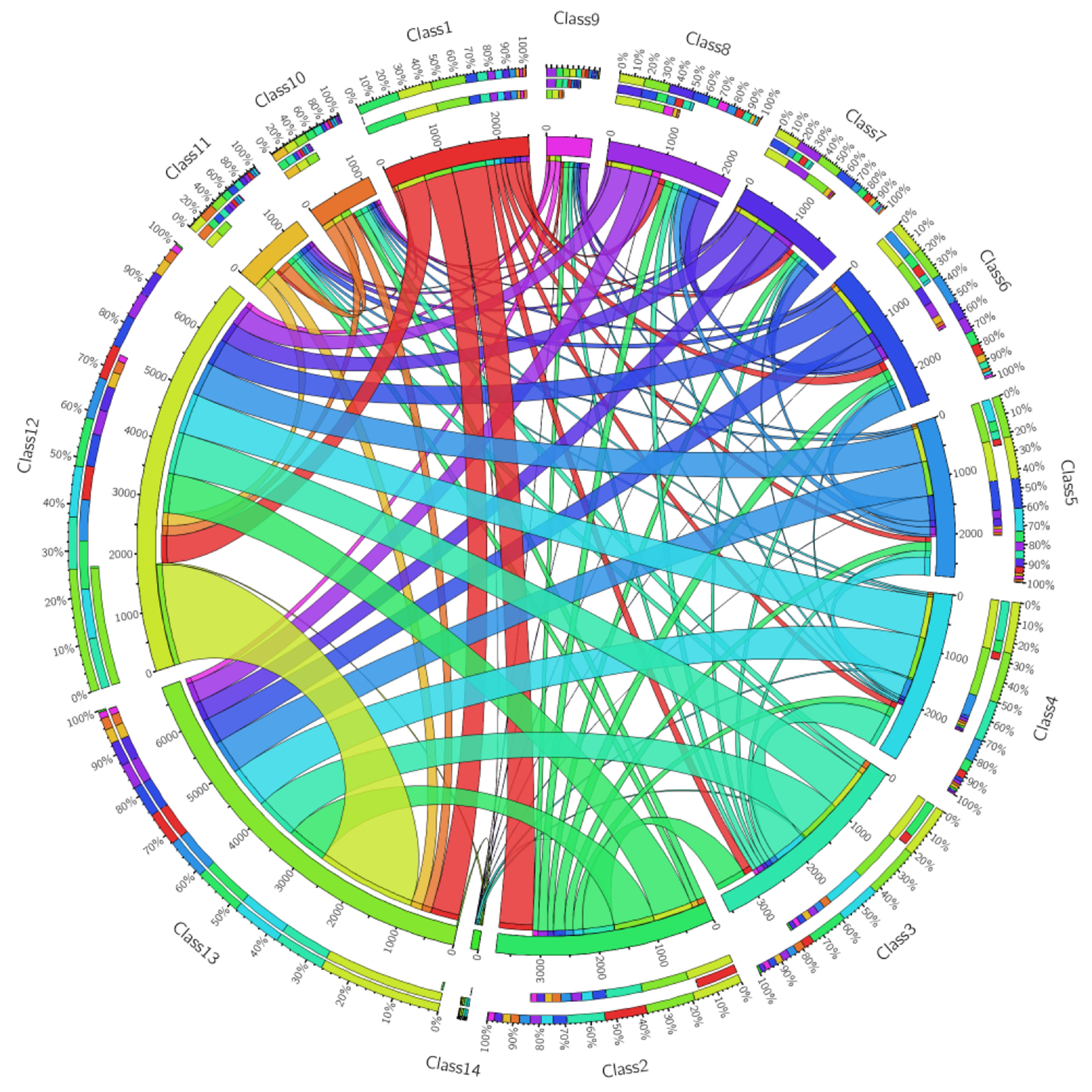}
\end{center}
\caption{Label concurrence in yeast MLD.} \label{yeast}
\end{figure*}

\subsection{Concurrence among Imbalanced Labels in MLDs}
The instances in a MLD are usually associated simultaneously to two or more labels. It is entirely possible that one of those labels is the minority label, while other is the majority one. In the most extreme situation, all the appearances of the minority label could be jointly with the majority one, into the same instances. This will make the minority label specially difficult to classify by any MLC algorithm, as most of them tend to be biased to the majority ones. In practice the scenario would be more complicated, as commonly there are more than one minority/majority label in an MLD. Therefore, the potential existence of instances  associated to minority and majority labels at once is very high. This fact is what we call concurrence among imbalanced labels.

A multilabel oversampling algorithm that clones minority labels, such as the proposed in \cite{Charte:Neucom13}, or that generates new samples from existing ones preserving the labelsets, as is the case in \cite{Giraldo:2013}, could be also increasing the number of instances associated to majority labels. Thus, the imbalance level would be hardly reduced if there is a high level of concurrence among imbalanced labels. In the same way, a multilabel undersampling algorithm designed to remove instances from the majority labels, such as the proposed in \cite{Charte:Neucom13}, could inadvertently cause also a loss of samples associated to the minority ones. In both cases, difficult labels (\textcolor {mycolor}{those} which are in minority and have a high concurrence with majority ones) will be the most harmed by the classifier.

The ineffectiveness of these resampling methods, when they are used with certain MLDs, would be noticed once the preprocessing is applied and the classification results are evaluated. This process will need computing power and time. For that reason, it would be desirable to know in advance the level of concurrence among imbalanced labels that each MLD suffers, saving these valuable resources.

\subsection{Metrics to assess the concurrence level}\label{SCUMBLE}
The concurrence of labels in an MLD can be visually explored in some cases, as shown in Figures \ref{genbase} and \ref{yeast} . Each arc represents a label, being the arc's length proportional to the number of instances in which this label is present. The first (Figure \ref{genbase}) diagram corresponds to the genbase dataset. At the position of twelve o'clock appears a label called \textit{P750} which is clearly a minority label. All the samples associated to this label also contains \textit{P271}, another minority label. The same situation can be seen with label \textit{P154}. These minority labels have not necessarily to be difficult labels.  By contrast, in the yeast MLD (Figure \ref{yeast}) is easy to see that the samples associated to minority labels, such as \textit{Class14} and \textit{Class9}, \textcolor {mycolor}{always appear} together with one or more majority labels. At first sight, that the concurrence between imbalanced labels is higher in yeast than in genbase, and that the former contains some difficult labels while the latter does not, could be concluded. However, this visual exploratory technique is not useful with MLDs having more than a few dozens labels, because the diagram would be hardly legible.

\textcolor {mycolor}{Existing metrics previously described (see Sect. \ref{ImbalanceMetrics}), such as \emph{IRLBL} and \emph{MeanIR}, assess the imbalance level of the labels, i. e. the relative frequency of each label with respect to the most common one and the average frequency. However, none of them allows to know if minority labels appear in their own or jointly with majority ones. The \emph{SCUMBLE} metric proposed here is aimed to evaluate this casuistic, that was not considered in the literature until now.}

The \textit{SCUMBLE} metric aims to quantify the imbalance variance among the labels present in each data sample. This metric (Eq. \ref{SCUMBLEEQ}) is based on the Atkinson index \cite{AtkinsonIndex} and the \textit{IRLbl} measure (Eq. \ref{IRLbl}) proposed in \cite{Charte:HAIS13}. The former is an econometric measure directed to assess social inequalities among individuals in a population. The latter is the \textcolor {mycolor}{metric} that lets us know the imbalance ratio of each label in an MLD. The Atkinson index is used to know the diversity among people's earnings, while our objective is to assess the extend to which labels with different imbalance levels appear jointly. Our first hypothesis is that the higher is the concurrence level the harder would be the work for resampling algorithms, and therefore the worse they would perform.

The Atkinson index is calculated using incomes\textcolor {mycolor}{. W}e used the imbalance level of each label instead, taking each instance $D_i$ in the MLD $D$ as a population, and the active labels in $D_{i}$ (those which are relevant to $D_i$ and therefore are set to 1) as the individuals. If the label $l$ is present in the instance $i$ then $IRLbl_{il} = IRLbl(l)$\textcolor {mycolor}{, otherwise }$IRLbl_{il} = 0$. $\overline{IRLbl_i}$ stands for the average imbalance level of the labels appearing in instance $i$. The scores for every sample are averaged, obtaining the final \textit{SCUMBLE} value.

\begin{equation}
  \textit{SCUMBLE}_{ins}\left(i\right) =
     1 - \frac{1}{\overline{\textit{IRLbl}_i}}\left(\prod\limits_{l=1}^{\mid{L}\mid} \textit{IRLbl}_{il}\right)^{\left(1/\mid{L}\mid\right)}
\label{SCUMBLEIns}
\end{equation}

\begin{equation}
  \textit{SCUMBLE}\left(D\right) = \frac{1}{\mid{D}\mid}
   \displaystyle\sum\limits_{i=1}^{\mid{D}\mid} \textit{SCUMBLE}_{ins}\left(i\right)
\label{SCUMBLEEQ}
\end{equation}

Since \textit{SCUMBLE} is computed as an average of concurrence by instance, it could be influenced by extreme values. A few instances with a very high \textit{SCUMBLE}$_{ins}$ value would introduce a certain deviation into the global \textit{SCUMBLE} measure. To estimate the importance of this deviation, the \textit{SCUMBLE.CV} metric (see Eq. \ref{SCUMBLE.CV}) provides the corresponding coefficient of variation. The higher is the \textit{SCUMBLE.CV}, the larger would be the differences in concurrence among instances.

\begin{equation}
\small
\begin{multlined} 
    \textit{SCUMBLE.CV} = \frac{\textit{SCUMBLE}\sigma}{\textit{SCUMBLE}}, \label{SCUMBLE.CV} \\ 
    \textit{SCUMBLE}\sigma = \sqrt{\displaystyle\sum\limits_{i=1}^{\mid D\mid}{\frac{\textit{(SCUMBLE}_{ins}(i) - \textit{SCUMBLE)}^2}{\mid D\mid-1}}}\\
\end{multlined}
\end{equation}

The \textit{SCUMBLE} measure for an MLD would provide a glimpse at how much concurrence between imbalanced labels there is in it. It also would be interesting to know which labels are more affected by this problem. This is the aim of the \textit{SCUMBLELbl} metric (Eq. \ref{SCUMBLELbl}). The \textit{SCUMBLELbl.CV} metric can also be obtained, following the same procedure described above for \textit{SCUMBLE.CV}. Since the number of instances in which the assessed label appears is used as denominator, dividing the sum of \textit{SCUMBLE}, that \textit{SCUMBLELbl} will be lower for majority labels is something intuitively deductible. Majority labels usually will interact with minority ones only in a few instances, those containing the minority label. Therefore, this metric would allow comparisons between labels with a similar frequency in the MLD. Our second hypothesis is that this information would be useful to know which of the minority labels are in fact heavily related to majority ones. In other words, which of them are difficult labels.

\begin{equation}
  \textit{SCUMBLELbl}\left(y\right) = \frac{\displaystyle\sum\limits_{i=1}^{\mid{D}\mid} \llbracket y \in Y_i \rrbracket . \textit{SCUMBLE}_{ins}\left(i\right)}{\displaystyle\sum\limits_{i=1}^{\mid D \mid} \llbracket  y \in Y_i \rrbracket}
\label{SCUMBLELbl}
\end{equation}

Whether our initial hypothesis are correct or wrong, and therefore these metrics are able to predict the difficulty that an MLD implies for resampling algorithms or not, is something to be proven experimentally.

\section{The Algorithm REMEDIAL}\label{REMEDIAL}
In this section the algorithm REMEDIAL, firstly introduced in \cite{Charte:REMEDIAL} as a specific method for MLDs with concurrence of highly imbalanced labels, is described. How REMEDIAL has been implemented into the mldr package, and how to use it, \textcolor {mycolor}{is also explained in Appendix \ref{mldrPackage}}.

As its name suggests, REMEDIAL (\textit{REsampling MultilabEl datasets by Decoupling highly ImbAlanced Labels}) is a method specifically designed for MLDs \textcolor {mycolor}{that} suffer from concurrence between imbalanced labels. In this context, \textit{highly imbalanced labels} has to be understood as labels with large differences in their \textit{IRLbls}. This is a fact assessed with the \textit{SCUMBLE} measure, thus REMEDIAL is directed to MLDs with a high \textit{SCUMBLE} level.

When the few samples in which a minority label is present also contain one or more majority labels, whose frequency in the MLD is much higher, the power of the input features to predict the labels might be biased to the majority ones. Our hypothesis is that, in a certain way, majority labels are masking the minority ones when they appear together, a problem that could be solved to some extent by decoupling the labels in these instances.

REMEDIAL is \textcolor {mycolor}{kind of} a resampling algorithm. It could be seen as an oversampling method, since it produces new instances in some cases. At the same time it also modifies existent samples. \textcolor {mycolor}{However, REMEDIAL never changes the number of samples associated to each label, i.e. the absolute frequency of the labels in the MLD.} In short, REMEDIAL is an editing plus oversampling algorithm, and it is an approach which has synergies with traditional resampling techniques.  The method pseudo-code is shown in Algorithm \ref{REMEDIALAlg}.

\algnewcommand{\LineComment}[1]{\State \(\triangleright\) #1}
\begin{algorithm} %
\caption{REMEDIAL algorithm.}
\label{REMEDIALAlg}
\begin{algorithmic}[1]
\Statex
\Function{REMEDIAL}{MLD $D$, Labels $L$}
    \LineComment{Calculate imbalance levels}
    \State \textit{IRLbl$_l$} $\gets$ calculateIRLbl($l$ in $L$) 
    \State \textit{IRMean} $\gets {\overline{IRLbl}}$
    \LineComment{Calculate SCUMBLE}
    \State \textit{SCUMBLEIns$_i$} $\gets$ calculateSCUMBLE($D_i$ in $D$) 
    \State \textit{SCUMBLE} $\gets {\overline{\textit{SCUMBLEIns}}}$
	\For{\textbf{each} \textit{instance i} \textbf{in} $D$}
      	\If{\textit{SCUMBLEIns$_i$} $> \textit{SCUMBLE}$ }
      	  \State $D'_i \gets D_i$ \Comment{Clone the affected instance}
      	  \LineComment{Maintain minority labels}
      	  \State $D_i[labels_{\textit{IRLbl} <= \textit{IRMean}}] \gets 0$ 
      	  \LineComment{Maintain majority labels}
      	  \State $D'_i[labels_{\textit{IRLbl} > \textit{IRMean}}] \gets 0$ 
      	  \State $D \gets D + D'_i$
      	\EndIf
	\EndFor

\EndFunction
\normalsize
\end{algorithmic}
\end{algorithm}

\renewcommand{\tabcolsep}{0.4em}
\begin{table*}
\begin{center}
{\color {mycolor}\begin{tabular}{lrrrrrrrrcc}
  \toprule
 \textbf{Dataset} & \textbf{Instances} & \textbf{Attributes} & \textbf{Labels} & \textbf{Labelsets} & \textbf{Card} & \textbf{Dens} & \textbf{MeanIR} & \textbf{MaxIR} & \textbf{SCUMBLE} & \textbf{Ref}. \\ 
  \midrule
corel5k     &  5 000 &    499 &   374 &  3 175 &  3.522 & 0.009 & 189.568 &1 120.000 & 0.394 & \cite{corel5k}\\ 
mediamill   & 43 907 &    120 &   101 &  6 555 &  4.376 & 0.043 & 256.405 &1 092.548 & 0.355 & \cite{mediamill}\\
cal500      &    502 &     68 &   174 &    502 & 26.044 & 0.150 &  20.578 &  88.800 & 0.337 & \cite{CAL500}\\
enron       &  1 702 &  1 001 &    53 &    753 &  3.378 & 0.064 &  73.953 & 913.000 & 0.303 & \cite{enron}\\ 
corel16k    & 13 618 &    500 &   144 &  4 692 &  2.815 & 0.020 &  32.998 & 116.407 & 0.279 & \cite{corel16k} \\ 
cs          &  9 270 &    635 &   274 &  4 749 &  2.556 & 0.009 &  85.002 & 226.700 & 0.272 & \cite{QUINTA} \\ 
tmc2007     & 28 596 & 49 060 &    22 &  1 341 &  2.158 & 0.098 &  15.157 &  41.980 & 0.175 & \cite{Srivastava:2005}\\
yeast       &  2 417 &    103 &    14 &    198 &  4.237 & 0.303 &   7.197 &  53.412 & 0.104 & \cite{Elisseeff1} \\ 
bibtex      &  7 395 &  1 836 &   159 &  2 856 &  2.402 & 0.015 &  12.498 &  20.431 & 0.094 & \cite{bibtex}\\
medical     &    978 &  1 449 &    45 &     94 &  1.245 & 0.028 &  89.501 & 266.000 & 0.047 & \cite{medical}\\
genbase     &    662 &  1 186 &    27 &     32 &  1.252 & 0.046 &  37.315 & 171.000 & 0.029 & \cite{genbase} \\ 
\bottomrule
\end{tabular}}
\caption{Main characteristics of the datasets.}
\label{TblDatasets}
\end{center}
\end{table*}

The \textit{IRLbl}, \textit{IRMean} and \textit{SCUMBLE} measures are computed in lines 2-7. \textit{SCUMBLE$_{Ins_i}$} is the concurrence level of the instance $D_i$. The mean \textit{SCUMBLE} for the MLD is obtained by averaging the individual \textit{SCUMBLE} for each sample.

Taking the mean \textit{SCUMBLE} as reference, only the samples with a \textit{SCUMBLEIns $>$ SCUMBLE} are processed. Those instances, which contain minority and majority labels, are decoupled into two instances, one containing only the majority labels and another one with the minority labels. In line 10 $D_i$, a sample affected by the problem at glance, is cloned in $D'_i$. The formula in line 12 edits the original $D_i$ instance by removing the majority labels from it. Majority labels are considered as those whose \textit{IRLbl} is equal or below to \textit{IRMean}. Line 14 does the opposite, removing from the cloned $D'_i$ the minority labels. $D_i$ belongs to the $D$ MLD, but $D'_i$ has to be added to it (line 15).

\textcolor {mycolor}{What differentiates REMEDIAL from existing resampling methods, such as the ones enumerated in Sect. \ref{ResamplingMethods}, is that it does not change the label frequencies in the MLD. All existent proposals increase the number of instances associated to minority labels or decrease the amount of samples linked to majority ones. On the other hand, the goal of REMEDIAL is to look for instances where minority and majority labels appear together, splitting them if is it necessary, but without deleting or adding labels. As far as we are concern, there is not a comparable method to REMEDIAL proposed in the literature.}

\section{Experimentation and Analysis}\label{Experimentation}
The conducted experimentation has been structured \textcolor {mycolor}{into} two phases. First, the interest is in checking how the \textit{SCUMBLE} level impacts the performance of some resampling methods. Second, how the proposed REMEDIAL algorithm influences the MLDs, and the classification behavior, is analyzed. The test bed framework is described in the next subsection, the obtained results and corresponding analysis of the two aforementioned phases are provided in the following ones.

\subsection{Experimental framework}

\textcolor{mycolor}{In the first phase of the experimentation, }to determine the usefulness of the \textit{SCUMBLE} metric, six of the MLDs shown in Table \ref{TblDatasets}, corel5k, cal500, enron, yeast, medical and genbase, were used. \textcolor {mycolor}{They have been chosen as representatives of different \textit{SCUMBLE} values, including the extreme levels,  corel5k (highest) and genbase (lowest), and four values which are in between.} The rightmost column indicates each dataset's origin. All of them are imbalanced, so theoretically they could benefit from \textcolor {mycolor}{applying} a resampling algorithm. Aside from the \textit{SCUMBLE} measure, the \textit{MaxIR} and \textit{MeanIR} values are also shown. \textcolor {mycolor}{These measurements correspond to whole datasets.} The values taken as reference point to the posterior analysis will be average values from training partitions\footnote[1]{The dataset partitions used in this experimentation, as well as color version of all figures, are available to download at \url{http://simidat.ujaen.es/SCUMBLE}.} using a  $2\times5$ folds scheme. The datasets appear in Table \ref{TblDatasets} sorted by \textit{SCUMBLE} value, from higher to lower. According to this measure, corel5k and cal500 would be among the most difficult MLDs \textcolor {mycolor}{in the first group}, since they have a high level of concurrence among labels with different imbalance levels. On the other hand, medical and genbase would be the most benefited from resampling\textcolor {mycolor}{, as most of the majority/minority labels in them do not appear together}.

Regarding the resampling algorithms, the two proposed in \cite{Charte:HAIS13} have been applied. Both are based on the LP transformation. LP-ROS does oversampling by cloning instances with minority labelsets, whereas LP-RUS performs undersampling removing samples associated to majority labelsets. All the dataset partitions were preprocessed, and the imbalance measures were calculated for each algorithm.

\textcolor{mycolor}{In the second phase of the experimentation, }to check the influence of REMEDIAL in classification results, the \textcolor {mycolor}{eleven MLDs shown in Table \ref{TblDatasets}} have been given as input, before and after preprocessing them with REMEDIAL, to \textcolor {mycolor}{six} different MLC algorithms:

\begin{itemize}
    \item BR (\textit{Binary Relevance}) \cite{Boutell}. Ensemble of binary classifiers. It is a transformation based method. A binary classifier is generated for each label and the individual predictions are joined to obtain the final prediction. 
    
    \item HOMER (\textit{Hierarchy of Multilabel Classifiers}) \cite{HOMER}. Ensemble of multiclass classifiers. It is a transformation method based on the label powerset approach, thus each label combination is interpreted as a label class. 
    
    \item IBLR (\textit{Instance-Based Learning by Logistic Regression}) \cite{Cheng}. Instance based classifier. IBLR is an improved version of ML-kNN \cite{Zhang1}, the best-known instance based multilabel classifier.
    
    {\color {mycolor}
    \item CLR (\textit{Calibrated Label Ranking}) \cite{CLR}. Ensemble of binary classifiers based on pair-wise comparisons. A binary classifier is generated for each label pair, instead of each label as in BR. The classifier produces a ranking of labels, from which the predicted labelset is obtained after applying a threshold.
    
    \item ECC (\textit{Ensemble of Classifier Chains}) \cite{Read}. Ensemble of binary classifiers based on chaining each model with the next one. The ensemble generates several chains setting the classifiers for each label at random locations in their respective chain.

    \item EPS (\textit{Ensemble of Pruned Sets}) \cite{Read:2008:2}. Ensemble of multiclass classifiers with pruned labelsets. Each classifier relies on the PS \cite{Read:2008} method to prune infrequent labelsets, easing the work of the underlying multiclass classifier.
    }
\end{itemize}

\textcolor {mycolor}{The C4.5 tree induction algorithm has been used as base classifier where an underlying binary or multiclass classifier is needed. Default parameters were used in all cases.}

\textcolor {mycolor}{As stated in \cite{Madjarov}, the performance of a multilabel classifier should be always assessed by means of several evaluation metrics. In this case,} classification results are evaluated using \textcolor {mycolor}{five} usual multilabel measures: Hamming Loss (HL), \textcolor {mycolor}{Precision}, Macro-FMeasure (MacroFM),  \textcolor {mycolor}{One Error (OE), and Ranking Loss (RL)}. HL (see Eq. \ref{HL}) is a global sample-based measure. It assesses differences between $Z_i$, the predicted labelset, and $Y_i$, the real one, without distinction among labels. The $\Delta$ operator returns the symmetric difference between both labelsets.  The lower the HL the better the predictions are. \textcolor {mycolor}{Precision (\ref{Precision}) is also example-based, and it is among the most usual performance metrics when it comes to evaluate a classifier.} MacroFM is the label-based version of the usual F-Measure (see Eq. \ref{Precision}, \ref{Recall} and \ref{F1}). As can be seen in Eq. \ref{MacroM}, in MacroFM F-Measure is evaluated independently for each label and then it is averaged. In the latter equation \textit{TP} stands for \textit{True Positives}, \textit{FP} for \textit{False Positives}, \textit{TN} for \textit{True Negatives}, and \textit{FN} for \textit{False Negatives}. \textcolor {mycolor}{OE (\ref{OneError}) and RL (\ref{RankingLoss}) are ranking-based evaluation metrics. In these equations, $rk(x_i,l)$ is a function that returns the confidence degree for the label $l$ in the prediction $Z_i$ provided by the classifier for the instance $x_i$. Additional information about all these metrics can be found in \cite{Charte:SB-MLC}.}

\begin{equation}\small
HL = \frac{1}{|D|} \displaystyle\sum\limits_{i=1}^{|D|} \frac{|Y_i \Delta Z_i|}{|L|} . \label{HL}
\end{equation}

\begin{equation}
Precision = \frac{1}{\mid D \mid} \displaystyle\sum\limits_{i=1}^{\mid D \mid} \frac{\mid Y_i \cap Z_i \mid }{\mid Z_i \mid } \label{Precision}
\end{equation}

\begin{equation}
Recall = \frac{1}{\mid D \mid } \displaystyle\sum\limits_{i=1}^{\mid D \mid } \frac{\mid Y_i \cap Z_i \mid}{\mid Y_i \mid} \label{Recall}
\end{equation}

\begin{equation}
\textit{F-Measure} = 2 * \frac{Precision * Recall}{Precision + Recall} \label{F1}
\end{equation}

\begin{equation}\small\label{MacroM}
\textit{MacroFM}=\frac{1}{|L|} \sum\limits_{i=1}^{|L|}\textit{F-Measure}(\textit{TP}_i,\textit{FP}_i,\textit{TN}_i,\textit{FN}_i)
\end{equation}

\newcommand{\opA}{\mathop{\vphantom{\sum}\mathchoice
    				{\vcenter{\hbox{\large argmax}}}
    				{\vcenter{\hbox{\large argmax}}}{\mathrm{argmax}}{\mathrm{argmax}}}\displaylimits}
{\color{mycolor}    				
\begin{equation}
\textit{OE} = \frac{1}{n} \displaystyle\sum\limits_{i=1}^{n} \llbracket [\opA\limits_{y \in Z_i} \langle rk(x_i, y) \rangle \notin Y_i] \rrbracket .\label{OneError}
\end{equation}

\begin{equation}
\begin{split}
\textit{RL} = \frac{1}{n} \displaystyle\sum\limits_{i=1}^{n}  \frac{1}{\lvert Y_i\rvert . \lvert\overline{Y_i}\rvert} 
		\lvert{y_a, y_b : rk(x_i, y_a) > rk(x_i, y_b)}\rvert, \\ (y_a, y_b) \in Y_i\times\overline{Y_i} 
\end{split}
\label{RankingLoss}
\end{equation}
}

\subsection{SCUMBLE influence in preprocessing and \\classification algorithms}
Once the LP-ROS and LP-RUS resampling algorithm\textcolor {mycolor}{s} were applied, the imbalance levels on the preprocessed MLDs were reevaluated. Table \ref{TblROS} shows the new \textit{MaxIR} and \textit{MeanIR} values for each dataset. Comparing these values with the \textcolor {mycolor}{original} ones, it can be verified that a general improvement in the imbalance levels has been achieved. Although there are some exceptions, in most cases both \textit{MaxIR} and \textit{MeanIR} are lower after applying the resampling algorithms.

\begin{table*}
\begin{center}
\begin{tabular}{lrrrr|rrrr}
  \toprule
  &  \multicolumn{4}{c|}{LP-ROS} & \multicolumn{4}{c}{LP-RUS} \\
   & \multicolumn{2}{c}{\textbf{MaxIR}} & \multicolumn{2}{c|}{\textbf{MeanIR}}  & \multicolumn{2}{c}{\textbf{MaxIR}} & \multicolumn{2}{c}{\textbf{MeanIR}} \\ 
   \textbf{Dataset} & \textbf{Before} & \textbf{After} & \textbf{Before} & \textbf{After}  & \textbf{Before} & \textbf{After} & \textbf{Before} & \textbf{After} \\ 
  \midrule
  corel5k & 896.000 & 969.400 & 166.057 & 140.743 & 896.000 & 817.100 & 166.057 & 155.032 \\ 
  cal500  & 133.192 & 179.358 &  21.274 &  25.468 & 133.192 & 133.192 &  21.274 &  21.274  \\ 
  enron   & 657.050 & 710.967 &  72.552 &  53.255 & 657.050 & 620.050 &  72.552 &  68.672 \\ 
  yeast   &  53.689 &  15.418 &   7.218 &   2.612 &  53.689 &  83.800 &   7.218 &  19.884 \\ 
  medical & 212.800 &  39.963 &  68.388 &  10.556 & 212.800 &  46.570 &  68.388 &   6.371 \\ 
  genbase & 136.800 &  13.703 &  31.665 &   4.500 & 136.800 & 150.800 &  31.665 &  51.157 \\ 
   \bottomrule
\end{tabular}
\caption{Imbalance levels after applying resampling algorithms (average values on training partitions).}
\label{TblROS}
\end{center}
\end{table*}

\begin{figure*}
\begin{center}
\fbox{\includegraphics[width=0.9\textwidth]{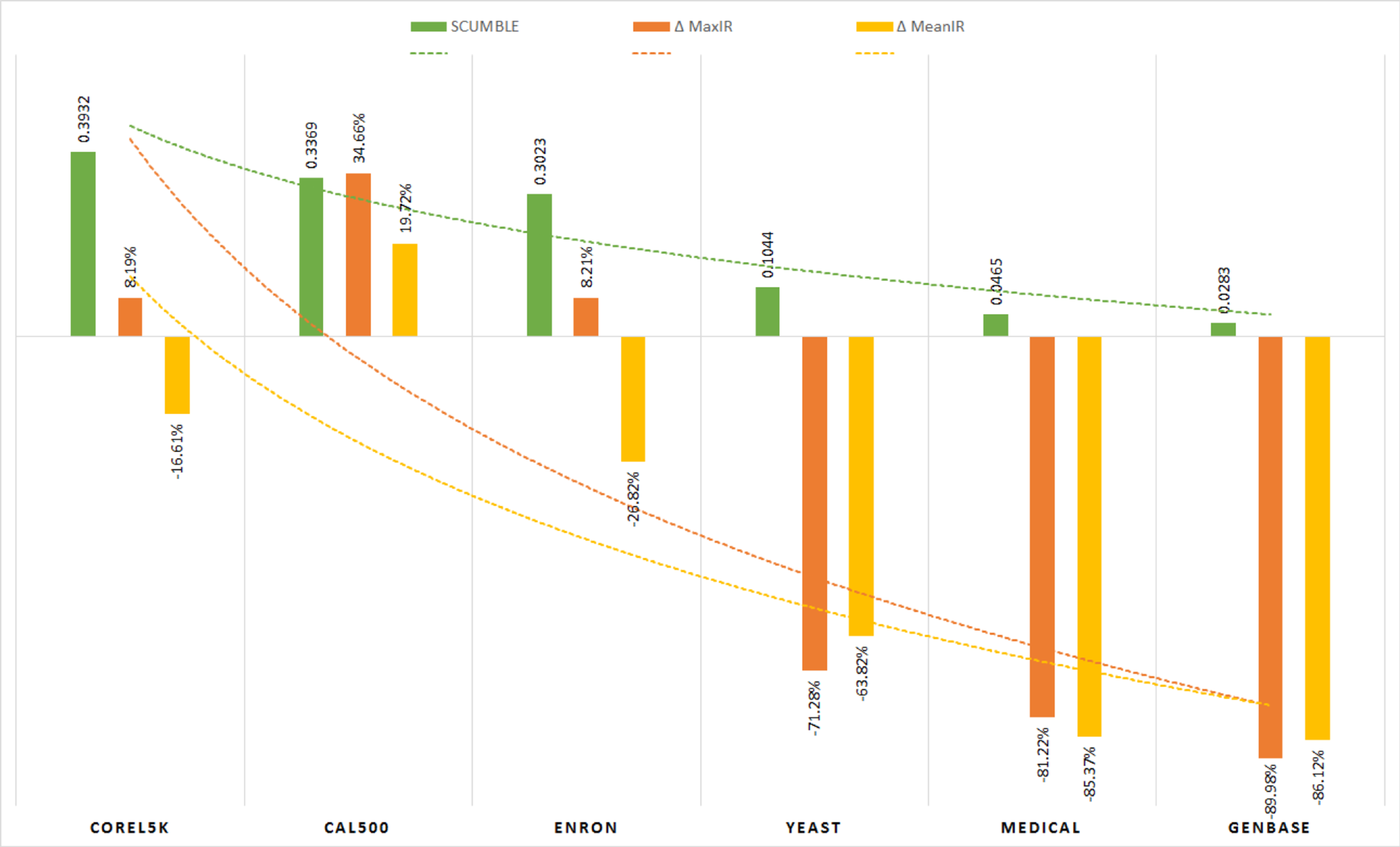}}
\end{center}
\caption{SCUMBLE vs changes in imbalance level after applying LP-ROS.} \label{LPROS}

\begin{center}
\fbox{\includegraphics[width=0.9\textwidth]{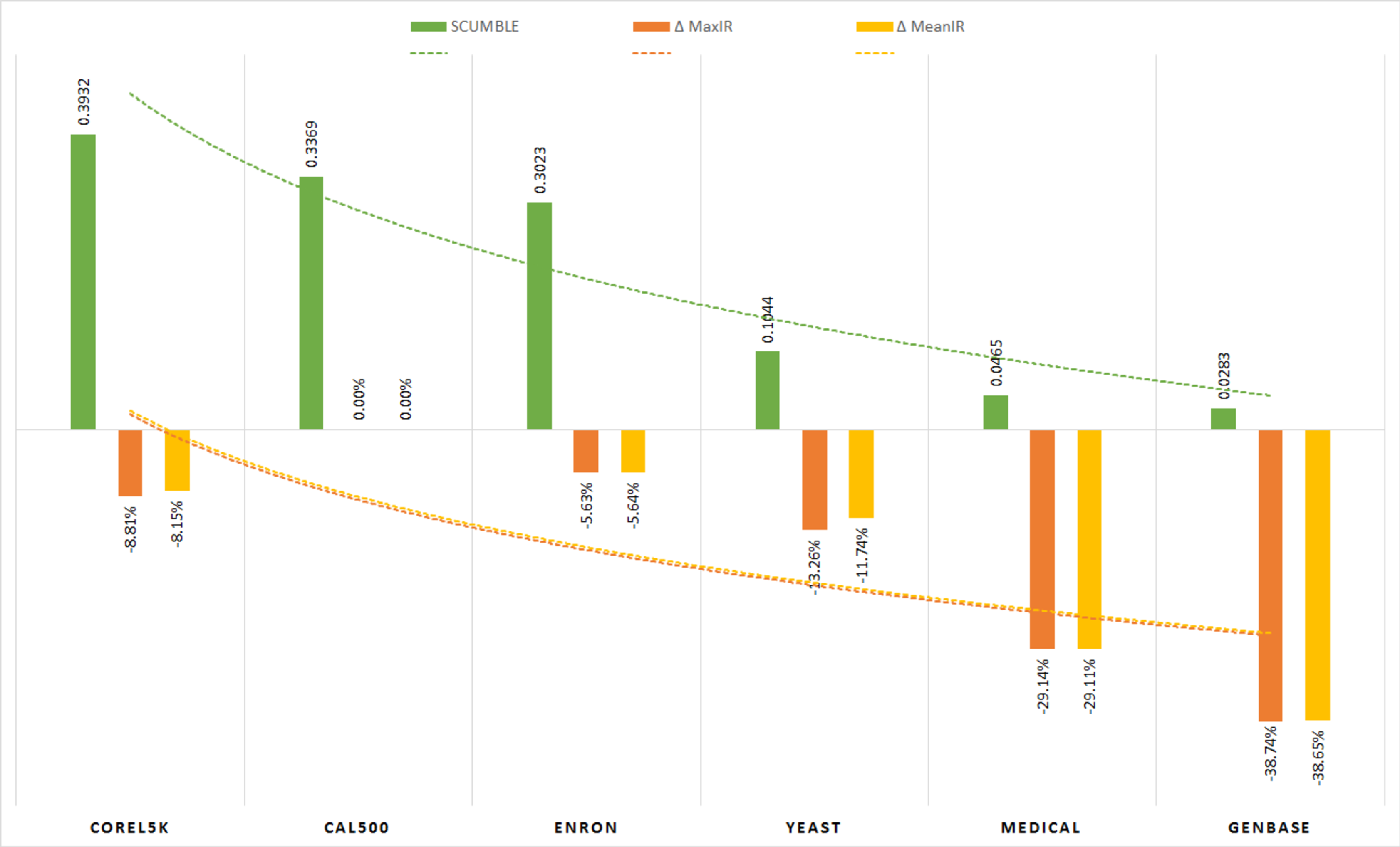}}
\end{center}
\caption{SCUMBLE vs changes in imbalance level after applying LP-RUS.} \label{LPRUS}
\end{figure*}

It would be interesting to know if the imbalance reduction is proportionally coherent with the values obtained from the \textit{SCUMBLE} measure. The graphs in Figure \ref{LPROS} and Figure \ref{LPRUS} are aimed to visually illustrate the connection between \textit{SCUMBLE} values and the relative variations in imbalance levels. For each MLD, their \textit{SCUMBLE} value is represented along with the percentage change in \textit{MaxIR} and \textit{MeanIR} after applying the LP-ROS/LP-RUS resampling methods. The tendency for the three values among the six MLDs is depicted by three logarithmic lines. As can be seen, a clear parallelism exists between the continuous line, which corresponds to \textit{SCUMBLE}, and the dashed lines. This affinity is specially remarkable with the LP-RUS algorithm (Figure \ref{LPRUS}).

Although the previous figures allow to infer that an important correlation between the \textit{SCUMBLE} measure and the success of the resampling algorithms exists, this relationship must be formally analyzed. To this end, a Pearson correlation test was applied over the \textit{SCUMBLE} values and the relative changes in imbalance levels for each resampling algorithm. The resulting correlation coefficients and \textit{p-values} are shown in Table \ref{TblCOR}. It can be seen that all the coefficients are above 80\%, and all the \textit{p-values} are under 0.05. Therefore, a statistical correlation between the \textit{SCUMBLE} measure and the behavior of the tested resampling algorithms can be concluded.

\renewcommand{\tabcolsep}{0.3em}
\begin{table}[h]
\begin{center}
\begin{tabular}{lcc|cc}
  \toprule
 &  \multicolumn{2}{c|}{\footnotesize SCUMBLE vs $\Delta$MaxIR} & \multicolumn{2}{c}{\footnotesize SCUMBLE vs $\Delta$MeanIR} \\
\textbf{Algorithm} & \textbf{Cor} & \textbf{p-value}  & \textbf{Cor} & \textbf{p-value} \\ 
  \midrule
  LP-ROS & 0.8120 & 0.0497  & 0.9189 & 0.0096 \\ 
  LP-RUS & 0.8607 & 0.0278  & 0.8517 & 0.0314 \\ 
   \bottomrule
\end{tabular}
\caption{Results from the Pearson correlation tests.}
\label{TblCOR}
\end{center}
\end{table}

Following this analysis, it seems reasonable to avoid resampling algorithms when the \textit{SCUMBLE} measure for an MLD is well above 0.1, such as is the case with corel5k, cal500 and enron. In this situation the benefits obtained from resampling, if any, are very small. The result can even be a worsening of the imbalance level. In average, the \textit{MeanIR} for the three MLDs with \textit{SCUMBLE} $> 0.3$ has been reduced only \textcolor {mycolor}{by} 6\%, while the \textit{MaxIR} is actually increasing in the same percentage. By contrast, the average \textit{MeanIR} reduction for the other three MLDs, with \textit{SCUMBLE} $\lesssim 0.1$, reaches 52\% and the \textit{MaxIR} reduction 54\%.

Aiming to know how these changes in the imbalance levels would influence classification results, and if a correlation with \textit{SCUMBLE} values exists, the HOMER \cite{HOMER} algorithm was used. It must be highlighted that the interest here is not in the raw performance values, but in how they change after a resampling algorithm has been applied and how this change correlates with \textit{SCUMBLE} values. Therefore, the HOMER algorithm is used only as a tool to obtain classification results before and after applying the resampling. Any other MLC algorithm could be used for this task. Additionally, the proposed \textit{SCUMBLE} measure is not used in the experimentation to influence the behavior of LP-ROS, LP-RUS or HOMER by any means. The goal is to explore the correlation between changes in classification results and \textit{SCUMBLE} values.

Table \ref{TblFMeasure} shows these results assessed with F-measure, the harmonic mean of precision and recall measures. It can be seen that with the three MLDs which show high \textit{SCUMBLE} values, the preprocessing has produced a remarkable deterioration in classification results. Among the other three MLDs the resampling has improved them in some cases, while producing a slight worsening (less than 1\%) in others. Therefore, even though the MLC algorithm behavior would be also affected by other dataset characteristics, that the \textit{SCUMBLE} \textcolor {mycolor}{metric} would offer valuable information to determine the convenience of applying a resampling method can be concluded.

\begin{table}
\begin{center}
\small
\begin{tabular}{lcccrr}
  \toprule
  \textbf{Dataset} & \textbf{Base} & \textbf{LP-RUS} & \textbf{LP-ROS} &  \textbf{$\Delta$RUS} &  \textbf{$\Delta$ROS} \\ 
  \midrule
  corel5k & 0.3857 & 0.2828 & 0.2920 & -26.6788 & -24.2935 \\ 
  cal500 & 0.3944 & 0.3127 & 0.3134 & -20.7150 & -20.5375 \\ 
  enron & 0.5992 & 0.5761 & 0.5874 & -3.8551 & -1.9693 \\ 
  yeast & 0.6071 & 0.6950 & 0.6966 & 14.4787 & 14.7422 \\ 
  medical & 0.9238 & 0.9158 & 0.9162 & -0.8660 & -0.8227 \\ 
  genbase & 0.9896 & 0.9818 & 0.9912 & -0.7882 & 0.1617 \\ 
   \bottomrule
\end{tabular}
\caption{F-Measure values obtained by HOMER MLC algorithm (average values over test partitions)}
\label{TblFMeasure}
\end{center}
\end{table}

\subsection{REMEDIAL experimental results}
Once the usefulness of the \textit{SCUMBLE} metric has been demonstrated, the next experimental phase has been applying the algorithm REMEDIAL to the \textcolor {mycolor}{eleven datasets previously shown in Table \ref{TblDatasets}}, then learning from them using \textcolor {mycolor}{six} multilabel classifiers. The results obtained from each one of them over the datasets, before and after preprocessing, are \textcolor {mycolor}{provided} in Table \ref{TblResultsHL} \textcolor {mycolor}{to Table \ref{TblResultsRL}. Each table correspond to one evaluation metric.} Best results are highlighted in bold. EPS was not able to process a couple of datasets.

\begin{table*}[ht!]
\centering
{\color {mycolor}
\begin{tabular}{lcc|cc|cc|cc|cc|cc}
\toprule
 & \multicolumn{2}{c}{BR} & \multicolumn{2}{c}{CLR} & \multicolumn{2}{c}{ECC} & \multicolumn{2}{c}{EPS} & \multicolumn{2}{c}{HOMER}  & \multicolumn{2}{c}{IBLR}\\
\cmidrule{2-13}
  \textbf{Dataset} & \textbf{Before} & \textbf{After} & \textbf{Before} & \textbf{After} & \textbf{Before} & \textbf{After} & \textbf{Before} & \textbf{After} & \textbf{Before} & \textbf{After} & \textbf{Before} & \textbf{After}\\ 
\midrule
  bibtex   & 0.0147 & \textbf{0.0132} & \textbf{0.0127} & 0.0130 & \textbf{0.0126} & 0.0135 & - & - & 0.0185 & \textbf{0.0166} & 0.0165 & \textbf{0.0155} \\ 
  cal500   & 0.1630 & \textbf{0.1497} & \textbf{0.1381} & 0.1442 & \textbf{0.1422} & 0.1490 & - & - & 0.1875 & \textbf{0.1815} & 0.2341 & \textbf{0.2125} \\ 
  corel16k & 0.0206 & \textbf{0.0196} & 0.0198 & \textbf{0.0195} & 0.0387 & \textbf{0.0195} & \textbf{0.0196} & 0.0268 & 0.0271 & \textbf{0.0228} & 0.0199 & \textbf{0.0198} \\ 
  corel5k  & 0.0098 & \textbf{0.0094} & 0.0095 & \textbf{0.0094} & 0.0094 & \textbf{0.0089} & 0.0173 & \textbf{0.0101} & 0.0132 & \textbf{0.0118} & 0.0242 & \textbf{0.0148} \\ 
  cs       & 0.0094 & \textbf{0.0089} & 0.0088 & 0.0088 & \textbf{0.0086} & 0.0578 & 0.0133 & \textbf{0.0112} & 0.0117 & \textbf{0.0104} & 0.0182 & \textbf{0.0143} \\ 
  enron    & \textbf{0.0522} & 0.0540 & \textbf{0.0476} & 0.0517 & 0.0484 & \textbf{0.0086} & 0.0733 & \textbf{0.0601} & 0.0574 & \textbf{0.0555} & \textbf{0.0571} & 0.0593 \\ 
  genbase  & \textbf{0.0012} & 0.0084 & \textbf{0.0014} & 0.0080 & \textbf{0.0014} & 0.0386 & \textbf{0.0028} & 0.0040 & \textbf{0.0016} & 0.0064 & \textbf{0.0022} & 0.0092 \\ 
  mediamill& 0.0343 & \textbf{0.0331} & \textbf{0.0291} & 0.0321 & 0.0288 & \textbf{0.0118} & 0.0524 & \textbf{0.0377} & 0.0384 & \textbf{0.0355} & \textbf{0.0291} & 0.0338 \\ 
  medical  & \textbf{0.0107} & 0.0131 & \textbf{0.0109} & 0.0132 & \textbf{0.0100} & 0.0717 & \textbf{0.0141} & 0.0143 & \textbf{0.0109} & 0.0118 & 0.0198 & 0.0198 \\ 
  tmc2007  & \textbf{0.0568} & 0.0684 & \textbf{0.0538} & 0.0658 & \textbf{0.0507} & 0.2316 & 0.0872 & \textbf{0.0693} & \textbf{0.0607} & 0.0647 & \textbf{0.0646} & 0.0775 \\ 
  yeast    & 0.2505 & \textbf{0.2347} & \textbf{0.2202} & 0.2228 & 0.3594 & \textbf{0.0094} & \textbf{0.2042} & 0.2853 & 0.2601 & \textbf{0.2476} & \textbf{0.1941} & 0.2264 \\ 
   \bottomrule
\end{tabular}
}
\caption{Results before and after applying REMEDIAL assessed with Hamming Loss ($\downarrow$)}\label {TblResultsHL}
\end{table*}

\begin{table*}[ht!]
\centering
{\color {mycolor}
\begin{tabular}{lcc|cc|cc|cc|cc|cc}
\toprule
 & \multicolumn{2}{c}{BR} & \multicolumn{2}{c}{CLR} & \multicolumn{2}{c}{ECC} & \multicolumn{2}{c}{EPS} & \multicolumn{2}{c}{HOMER}  & \multicolumn{2}{c}{IBLR}\\
\cmidrule{2-13}
  \textbf{Dataset} & \textbf{Before} & \textbf{After} & \textbf{Before} & \textbf{After} & \textbf{Before} & \textbf{After} & \textbf{Before} & \textbf{After} & \textbf{Before} & \textbf{After} & \textbf{Before} & \textbf{After}\\ 
\midrule
bibtex   & 0.5770 & \textbf{0.7451} & 0.8267 & \textbf{0.9173} & 0.7385 & \textbf{0.8023} & -      & -      & 0.4701 & \textbf{0.5356} & 0.4920 & \textbf{0.5004} \\ 
cal500   & 0.4397 & \textbf{0.5326} & 0.6363 & \textbf{0.8327} & 0.5636 & \textbf{0.6848} & -      & -      & 0.3842 & \textbf{0.3983} & \textbf{0.2859} & 0.2743 \\ 
corel16k & 0.3610 & \textbf{0.4682} & 0.4455 & \textbf{0.6056} & 0.1944 & \textbf{0.6913} & \textbf{0.4697} & 0.2010 & 0.2475 & \textbf{0.2921} & \textbf{0.3623} & 0.3079 \\ 
corel5k  & 0.3643 & \textbf{0.4781} & 0.4621 & \textbf{0.5983} & 0.5465 & \textbf{0.6868} & 0.1938 & \textbf{0.2906} & 0.2232 & \textbf{0.2438} & 0.0598 & \textbf{0.0602} \\ 
cs       & 0.5174 & \textbf{0.6239} & 0.6297 & \textbf{0.7247} & 0.6211 & \textbf{0.7611} & 0.3366 & \textbf{0.3632} & 0.3884 & \textbf{0.4394} & \textbf{0.1076} & 0.0919 \\ 
enron    & 0.6391 & \textbf{0.7063} & 0.7047 & \textbf{0.7813} & 0.6681 & \textbf{0.9960} & 0.4873 & \textbf{0.5592} & 0.5893 & \textbf{0.6269} & 0.6151 & \textbf{0.6519} \\ 
genbase  & 0.9947 & \textbf{0.9977} & 0.9946 & \textbf{0.9977} & \textbf{0.9950} & 0.8877 & \textbf{0.9950} & 0.9942 & 0.9932 & \textbf{0.9961} & \textbf{0.9899} & 0.9890 \\ 
mediamill& 0.6683 & \textbf{0.8091} & 0.7959 & \textbf{0.8707} & 0.7986 & \textbf{0.8740} & 0.4827 & \textbf{0.6271} & 0.6177 & \textbf{0.6805} & 0.7758 & \textbf{0.8388} \\ 
medical  & 0.8633 & \textbf{0.8680} & 0.8699 & \textbf{0.8725} & 0.8636 & \textbf{0.8701} & 0.7813 & \textbf{0.7837} & 0.8639 & \textbf{0.8648} & 0.7272 & \textbf{0.7552} \\ 
tmc2007  & 0.7675 & \textbf{0.8334} & 0.7855 & \textbf{0.8539} & \textbf{0.8056} & 0.7370 & 0.6079 & \textbf{0.6903} & 0.7389 & \textbf{0.7829} & 0.7309 & \textbf{0.8031} \\ 
yeast    & 0.6020 & \textbf{0.6647} & 0.6768 & \textbf{0.7323} & 0.4777 & \textbf{0.6899} & \textbf{0.6960} & 0.5422 & 0.5876 & \textbf{0.6169} & 0.7110 & \textbf{0.7442} \\ 
   \bottomrule
\end{tabular}
}
\caption{Results before and after applying REMEDIAL assessed with Precision ($\uparrow$)}\label {TblResultsPrecision}
\end{table*}

\begin{table*}[ht!]
\centering
{\color {mycolor}
\begin{tabular}{lcc|cc|cc|cc|cc|cc}
\toprule
 & \multicolumn{2}{c}{BR} & \multicolumn{2}{c}{CLR} & \multicolumn{2}{c}{ECC} & \multicolumn{2}{c}{EPS} & \multicolumn{2}{c}{HOMER}  & \multicolumn{2}{c}{IBLR}\\
\cmidrule{2-13}
  \textbf{Dataset} & \textbf{Before} & \textbf{After} & \textbf{Before} & \textbf{After} & \textbf{Before} & \textbf{After} & \textbf{Before} & \textbf{After} & \textbf{Before} & \textbf{After} & \textbf{Before} & \textbf{After}\\ 
\midrule
bibtex   & 0.3368 & \textbf{0.3604} & 0.3342 & \textbf{0.3518} & 0.3750 & \textbf{0.3763} & -      & -      & 0.2984 & \textbf{0.2985} & \textbf{0.2140} & 0.1950 \\ 
cal500   & \textbf{0.2933} & 0.2286 & \textbf{0.3323} & 0.2436 & \textbf{0.3058} & 0.0670 & -      & -      & 0.3301 & \textbf{0.3372} & \textbf{0.2772} & 0.2527 \\ 
corel16k & \textbf{0.1550} & 0.1266 & \textbf{0.1084} & 0.0707 & \textbf{0.1477} & 0.1056 & 0.1223 & \textbf{0.1401} & \textbf{0.1510} & 0.1377 & \textbf{0.1146} & 0.0956 \\ 
corel5k  & 0.1774 & \textbf{0.1827} & \textbf{0.1330} & 0.1073 & 0.1666 & \textbf{0.2922} & \textbf{0.1860} & 0.1767 & \textbf{0.1963} & 0.1860 & 0.1060 & \textbf{0.1432} \\ 
cs       & 0.3457 & \textbf{0.3795} & \textbf{0.2801} & 0.2606 & \textbf{0.3617} & 0.2760 & \textbf{0.3044} & 0.2992 & \textbf{0.2999} & 0.2922 & \textbf{0.1355} & 0.1341 \\ 
enron    & 0.4029 & \textbf{0.4189} & \textbf{0.4199} & 0.3755 & 0.4324 & \textbf{0.8970} & 0.3828 & \textbf{0.3933} & \textbf{0.3836} & 0.3828 & \textbf{0.3458} & 0.2755 \\ 
genbase  & 0.9890 & \textbf{0.9923} & \textbf{0.9848} & 0.9415 & \textbf{0.9906} & 0.1741 & \textbf{0.9775} & 0.9527 & \textbf{0.9806} & 0.9662 & \textbf{0.9655} & 0.8449 \\ 
mediamill& 0.2836 & \textbf{0.2959} & \textbf{0.2307} & 0.2011 & 0.2445 & \textbf{0.8085} & \textbf{0.3382} & 0.2657 & \textbf{0.2492} & 0.2147 & \textbf{0.2818} & 0.1820 \\ 
medical  & \textbf{0.8165} & 0.8013 & \textbf{0.7942} & 0.7864 & \textbf{0.8179} & 0.3318 & 0.7283 & \textbf{0.7292} & \textbf{0.7981} & 0.7855 & \textbf{0.6404} & 0.6189 \\ 
tmc2007  & \textbf{0.6015} & 0.4243 & \textbf{0.6073} & 0.3578 & \textbf{0.5966} & 0.4063 & 0.5802 & \textbf{0.5951} & \textbf{0.5981} & 0.4551 & \textbf{0.4667} & 0.2786 \\ 
yeast    & 0.4341 & \textbf{0.5204} & \textbf{0.4480} & 0.4075 & \textbf{0.4782} & 0.1356 & \textbf{0.4629} & 0.4428 & 0.4363 & \textbf{0.4400} & \textbf{0.4945} & 0.3901 \\ 
   \bottomrule
\end{tabular}
}
\caption{Results before and after applying REMEDIAL assessed with Macro F-Measure ($\uparrow$)}\label {TblResultsMacroFM}
\end{table*}

\begin{table*}[ht!]
\centering
{\color {mycolor}
\begin{tabular}{lcc|cc|cc|cc|cc|cc}
\toprule
 & \multicolumn{2}{c}{BR} & \multicolumn{2}{c}{CLR} & \multicolumn{2}{c}{ECC} & \multicolumn{2}{c}{EPS} & \multicolumn{2}{c}{HOMER}  & \multicolumn{2}{c}{IBLR}\\
\cmidrule{2-13}
  \textbf{Dataset} & \textbf{Before} & \textbf{After} & \textbf{Before} & \textbf{After} & \textbf{Before} & \textbf{After} & \textbf{Before} & \textbf{After} & \textbf{Before} & \textbf{After} & \textbf{Before} & \textbf{After}\\ 
\midrule
bibtex   & 0.5060 & \textbf{0.4648} & \textbf{0.4110} & 0.4120 & \textbf{0.3886} & 0.4018 & -      &  -     & 0.6040 & 0.6040 & \textbf{0.6043} & 0.6651 \\ 
cal500   & 0.7202 & \textbf{0.6963} & 0.1254 & \textbf{0.1245} & \textbf{0.1504} & 0.3735 & -      &  -     & 0.8167 & \textbf{0.7660} & \textbf{0.8756} & 0.8885 \\ 
corel16k & \textbf{0.6964} & 0.7289 & \textbf{0.6650} & 0.6696 & 0.7138 & \textbf{0.7036} & \textbf{0.6809} & 0.7906 & \textbf{0.7712} & 0.8065 & \textbf{0.7106} & 0.7359 \\ 
corel5k  & \textbf{0.7067} & 0.7134 & \textbf{0.6716} & 0.6753 & 0.6828 & \textbf{0.5181} & \textbf{0.7853} & 0.9070 & \textbf{0.7994} & 0.8165 & 0.9401 & \textbf{0.9066} \\ 
cs       & 0.5690 & \textbf{0.5679} & \textbf{0.5112} & 0.5130 & 0.4833 & \textbf{0.3144} & \textbf{0.5596} & 0.6629 & \textbf{0.6680} & 0.7042 & 0.9041 & \textbf{0.8711} \\ 
enron    & 0.3922 & \textbf{0.3554} & 0.2350 & \textbf{0.2311} & 0.2700 & \textbf{0.0037} & \textbf{0.3044} & 0.4168 & 0.4456 & \textbf{0.4360} & \textbf{0.3805} & 0.3734 \\ 
genbase  & \textbf{0.0052} & 0.0060 & \textbf{0.0022} & 0.0030 & \textbf{0.0022} & 0.1769 & \textbf{0.0037} & 0.0068 & 0.0114 & \textbf{0.0106} & \textbf{0.0098} & 0.0384 \\ 
mediamill& 0.3943 & \textbf{0.2093} & \textbf{0.1125} & 0.1155 & \textbf{0.1153} & 0.1616 & \textbf{0.1155} & 0.1651 & 0.3839 & \textbf{0.3479} & \textbf{0.1215} & 0.1351 \\ 
medical  & \textbf{0.1906} & 0.1984 & 0.1559 & \textbf{0.1544} & \textbf{0.1534} & 0.1716 & \textbf{0.1830} & 0.1917 & \textbf{0.2107} & 0.2224 & 0.3190 & \textbf{0.3175} \\ 
tmc2007  & 0.2374 & \textbf{0.2137} & 0.1575 & \textbf{0.1544} & \textbf{0.1603} & 0.2685 & \textbf{0.1855} & 0.2122 & \textbf{0.2788} & 0.2836 & \textbf{0.2298} & 0.2345 \\ 
yeast    & 0.4181 & \textbf{0.3856} & 0.2399 & \textbf{0.2366} & \textbf{0.2574} & 0.7249 & \textbf{0.2520} & 0.2873 & 0.4268 & \textbf{0.3885} & \textbf{0.2255} & 0.2441 \\ 
   \bottomrule
\end{tabular}
}
\caption{Results before and after applying REMEDIAL assessed with One Error ($\downarrow$)}\label {TblResultsOE}
\end{table*}

\begin{table*}[ht!]
\centering
{\color {mycolor}
\begin{tabular}{lcc|cc|cc|cc|cc|cc}
\toprule
 & \multicolumn{2}{c}{BR} & \multicolumn{2}{c}{CLR} & \multicolumn{2}{c}{ECC} & \multicolumn{2}{c}{EPS} & \multicolumn{2}{c}{HOMER}  & \multicolumn{2}{c}{IBLR}\\
\cmidrule{2-13}
  \textbf{Dataset} & \textbf{Before} & \textbf{After} & \textbf{Before} & \textbf{After} & \textbf{Before} & \textbf{After} & \textbf{Before} & \textbf{After} & \textbf{Before} & \textbf{After} & \textbf{Before} & \textbf{After}\\ 
\midrule
bibtex   & \textbf{0.1635} & 0.1986 & \textbf{0.0620} & 0.0622 & \textbf{0.0950} & 0.1206 & -      & -      & \textbf{0.3295} & 0.3382 & \textbf{0.1742} & 0.1824 \\ 
cal500   & 0.3159 & \textbf{0.2086} & 0.1809 & \textbf{0.1805} & \textbf{0.1975} & 0.2343 & -      & -      & 0.3911 & \textbf{0.3775} & 0.3790 & \textbf{0.3226} \\ 
corel16k & 0.1857 & \textbf{0.1848} & 0.1335 & 0.1335 & 0.2907 & \textbf{0.1840} & \textbf{0.1772} & 0.3367 & \textbf{0.3973} & 0.4228 & \textbf{0.1687} & 0.1733 \\ 
corel5k  & 0.1474 & \textbf{0.1416} & 0.1176 & 0.1176 & 0.1414 & \textbf{0.1295} & \textbf{0.4807} & 0.6748 & \textbf{0.4387} & 0.4690 & 0.2754 & 0.\textbf{2620} \\ 
cs       & 0.1996 & \textbf{0.1703} & 0.0672 & 0.0672 & 0.1118 & \textbf{0.0911} & \textbf{0.2604} & 0.3409 & \textbf{0.3532} & 0.3770 & \textbf{0.2234} & 0.2271 \\ 
enron    & 0.1746 & \textbf{0.1409} & 0.0737 & \textbf{0.0736} & 0.0852 & \textbf{0.0062} & \textbf{0.1667} & 0.2333 & \textbf{0.2502} & 0.2759 & 0.1066 & 0.1066 \\ 
genbase  & \textbf{0.0030} & 0.0137 & \textbf{0.0088} & 0.0089 & \textbf{0.0036} & 0.0534 & \textbf{0.0078} & 0.0096 & \textbf{0.0060} & 0.0240 & \textbf{0.0040} & 0.0326 \\ 
mediamill& 0.1742 & \textbf{0.0761} & \textbf{0.0336} & 0.0338 & 0.0439 & \textbf{0.0394} & \textbf{0.0738} & 0.1154 & \textbf{0.2162} & 0.2236 & \textbf{0.0391} & 0.0404 \\ 
medical  & \textbf{0.0703} & 0.0785 & 0.0297 & 0.0297 & \textbf{0.0357} & 0.0547 & 0.0686 & \textbf{0.0664} & \textbf{0.0999} & 0.1045 & 0.0653 & \textbf{0.0640} \\ 
tmc2007  & \textbf{0.1139} & 0.1271 & 0.0347 & \textbf{0.0338} & \textbf{0.0444} & 0.1895 & \textbf{0.0631} & 0.0956 & \textbf{0.1547} & 0.1852 & \textbf{0.0558} & 0.0589 \\ 
yeast    & 0.3156 & \textbf{0.2536} & 0.1799 & \textbf{0.1785} & \textbf{0.2021} & 0.1442 & \textbf{0.1854} & 0.2241 & 0.3407 & \textbf{0.3213} & \textbf{0.1643} & 0.1768 \\ 
   \bottomrule
\end{tabular}
}
\caption{Results before and after applying REMEDIAL assessed with Ranking Loss ($\downarrow$)}\label {TblResultsRL}
\end{table*}

\begin{table*}[ht]
\renewcommand{\tabcolsep}{1em}
\centering
{\color {mycolor}
\begin{tabular}{lrrrrr}
  \toprule
Classifier & Hamming Loss & Precision & Macro F-Measure & One Error & Ranking Loss \\ 
  \midrule
BR    & 0.824098 & $^*$0.003857 & 0.893904 & 0.142368 & 0.266402 \\ 
CLR   & $^*$0.016316 & $^*$0.003857 & $^*$0.006692 & 0.893904 & 0.824098 \\ 
ECC   & 0.689084 & $^*$0.029383 & 0.398305 & 0.449804 & 0.964541 \\ 
EPS   & 0.150786 & 0.352542 & 0.932647 & 0.150786 & 0.150786 \\ 
HOMER & 0.168167 & $^*$0.003857 & $^*$0.029383 & 0.893904 & $^*$0.045447 \\ 
IBLR  & 0.683481 & 0.266402 & $^*$0.018408 & 0.398305 & 0.414823 \\ 
   \bottomrule
\end{tabular}
}
\caption{Exact p-values produced by the Wilcoxon statistical test for each classifier/metric.}\label{TblWilcoxon}
\end{table*}

{\color {mycolor}
The analysis of these results can be structured into three parts depending on where we put the focus, the classifiers, the datasets or the evaluation metrics.
\begin{itemize}
    % BR: 39/16, HOMER: 29/26, ECC: 28/27, EPS: 16/19, CLR: 23/32, IBLR: 22/33
    \item Going through the results by classifier, that REMEDIAL works better with BR and HOMER than with IBLR and CLR can be easily observed. The results for ECC and EPS are not conclusive, with almost as many cases with improvements and worsenings. Binary relevance based algorithms train a classifier for each label, taking as positive the instances containing it and as negative the remainder samples. When a majority label is being processed, all the instances in which it appears jointly with a minority label are processed as positive, disregarding the fact that they contain other labels. The decoupling of these labels tends to balance the bias of each classifier, something that also influences the behavior of ECC. LP based algorithms, such as HOMER, surely are favored by REMEDIAL, since the decoupling produces simpler labelsets. Moreover, the number of distinct labelsets is reduced after the resampling. The influence of REMEDIAL on instance based classifiers, such as IBLR, is easy to devise. The attributes of the decoupled samples do not change, so they will occupy exactly the same position with respect to the instance which is taken as reference for searching nearest neighbors. Therefore, the classifier will get two samples at the same distance but with disjoint labelsets, something that can be confusing depending on how the algorithm predicts the labelset of the reference sample.
    
    % enron: 21/9, corel5k: 20/10, cal500: 13/7, yeast: 17/13, mediamill: 16/14, bibtext: 11/9, cs: 15/15, corel16k: 13/17, medical: 12/18, tcm2007: 10/20, genbase: 5/25
    \item Analyzing the results by dataset, two thirds of the best values for enron, corel5k and cal500 are obtained after applying REMEDIAL. As can be checked, these are the datasets with highest \textit{SCUMBLE} levels. On the other hand, the results that correspond to the genbase, medical and tmc2007 have not improvements. As shown in Table \ref{TblDatasets}, these are three datasets with low \textit{SCUMBLE} values. Although some differences are quite small, in general the decoupling of labels has worsened classification performance. The remainder five MLDs get mixed results, although this trend (the higher the \textit{SCUMBLE} level the more the result is improved) is similar. As a consequence a clear guideline follows from the analysis of these results, REMEDIAL should not be used with MLDs with low \textit{SCUMBLE} levels, since it is an algorithm specifically designed to face the opposite casuistic.
    
    % HL: 33/29, Prec: 11/51, FM: 34/28, OE: 31/31, RL: 33/29
    \item Lastly, focusing on the evaluation metrics, that Precision is higher after applying REMEDIAL for most of the datasets and classifiers, with only 9 out of 64 (14\%) cases without improvements, can be observed. According to the other four evaluation metrics, HL, MacroFM, OE and RL, there is almost a tie between cases whose results have been improved and those which have not achieved this goal. The view changes drastically depending on each classifier/metric combination. For instance, the HL and Precision values for HOMER state that REMEDIAL improves results in 19 out of 22 cases (86\%), but MacroFM and RL indicates the same only for 5 out of 22 (23\%).
\end{itemize}
}

{\color {mycolor}The statistical significance of the differences in the results just pointed out has been assessed by means of a paired Wilcoxon statistical test. The exact p-values for each metric/classifier are the shown in Table \ref{TblWilcoxon}. Those preceded with a $^*$ symbol can be considered as significant from a statistical point of view, applying the usual 0.05 threshold. Most of the differences are not statistically significant. However, Precision and MacroFM show important differences in half or more of the cases. The former metric reveals statistically significant improvements with BR, CLR, ECC and HOMER. On the contrary, MacroFM indicates that the worsening of results is remarkable for CLR, HOMER and IBLR.}

Overall, REMEDIAL would be a recommended resampling for MLDs with high \textit{SCUMBLE} levels and when BR or LP based classifiers are going to be used. In these cases the prediction of minority labels would be improved, and the global performance of the classifiers would be better. {\color {mycolor}MLDs such as genbase, medical and tmc2007, as their intrinsic traits have demonstrated, should not be processed with REMEDIAL. The same would be applicable to classifiers such as IBLR, as putting two data samples at the same location but having disjoint labelsets tend to confuse this kind of algorithms. Excluding these cases, the global evaluation of the results produced by REMEDIAL would be much more positive.

The already described before are the benefits brought by REMEDIAL on their own, but this algorithm could be used as a first step aimed to ease the work of traditional resampling techniques. Mixing REMEDIAL with standard oversampling and undersampling techniques would be an interesting further study.}

\section{Conclusions}\label{Conclusions}
From the conducted experimentation and further analysis it can be inferred that, while working with imbalanced MLDs, standard resampling methods should be avoided when the \textit{SCUMBLE} level is well above 0.1. In this situation the benefits from resampling are almost negligible, or even detrimental.

In the described scenario, with MLDs suffering from high concurrence among imbalance labels, the proposed REMEDIAL algorithm has proven to be effective. The algorithm looks for instances with a high \textit{SCUMBLE} level and decouples minority and majority labels, producing new instances. The conducted experimentation has proven that REMEDIAL is able to improve classification results when applied to MLDs with a high \textit{SCUMBLE}.

How to assess the concurrence problem, and how to deal with it in practice, has been explained by means of the mldr R package \textcolor {mycolor}{(see Appendix \ref{mldrPackage})}. This software has been extended by the authors to include the metrics and algorithms described in this paper. The goal is to help anyone interested in this topic to conduct their personal analysis.

It could be concluded that basic resampling algorithms, which clone the labelsets in new instances or remove samples, are not a general solution in the multilabel field. More sophisticated approaches, which take into account the concurrence among imbalanced labels, would be needed. A potential way for designing these new algorithms would be joining REMEDIAL with some of the existent resampling methods. Once the labels have been decoupled, traditional oversampling and undersampling algorithms would find less obstacles to do their work.

\textbf{Acknowledgments}: This work is partially supported by the Spanish Ministry of Science and Technology under projects TIN2014-57251-P and TIN2012-33856, and the Andalusian regional project P11-TIC-7765. 

\section*{References}

%\bibliographystyle{elsarticle-num}
%\bibliography{fcharte}

\newpage
\appendix
\renewcommand*{\thesection}{\Alph{section}}
\section{SCUMBLE and REMEDIAL implementations in the mldr package}\label{mldrPackage}
This appendix describes how to obtain the \textit{SCUMBLE} measurement for any MLD, as well as how to apply the REMEDIAL algorithm proposed in this study to any MLD, by means of a software package developed by the same authors.

\subsection{Assessing label concurrence with the mldr package}
The mldr package \cite{Charte:mldr} provides an easy way to make exploratory analysis over MLDs from R, one of the best known tools for machine learning tasks. A quick tutorial describing how to install and use this package can be found in \cite{Charte:BasicMLDR}. The capabilities of the mldr package \textcolor {mycolor}{have} been extended to include functions aimed to ease the concurrence analysis in MLDs. These new capabilities, developed ad hoc for the present work, are described below.

Once the package has been loaded into R, the first step will be reading the MLD to analyze. MULAN \cite{MULAN} and MEKA \cite{MEKA} file formats are supported. In order to load an MLD, the \texttt{mldr} function has to be called providing the file name. The returned result is an S3 R object containing the data (instances with attribute values) and also a plethora of characterization metrics. 

\begin{figure*}[ht!]
\begin{center}
  \includegraphics[width=.8\textwidth]{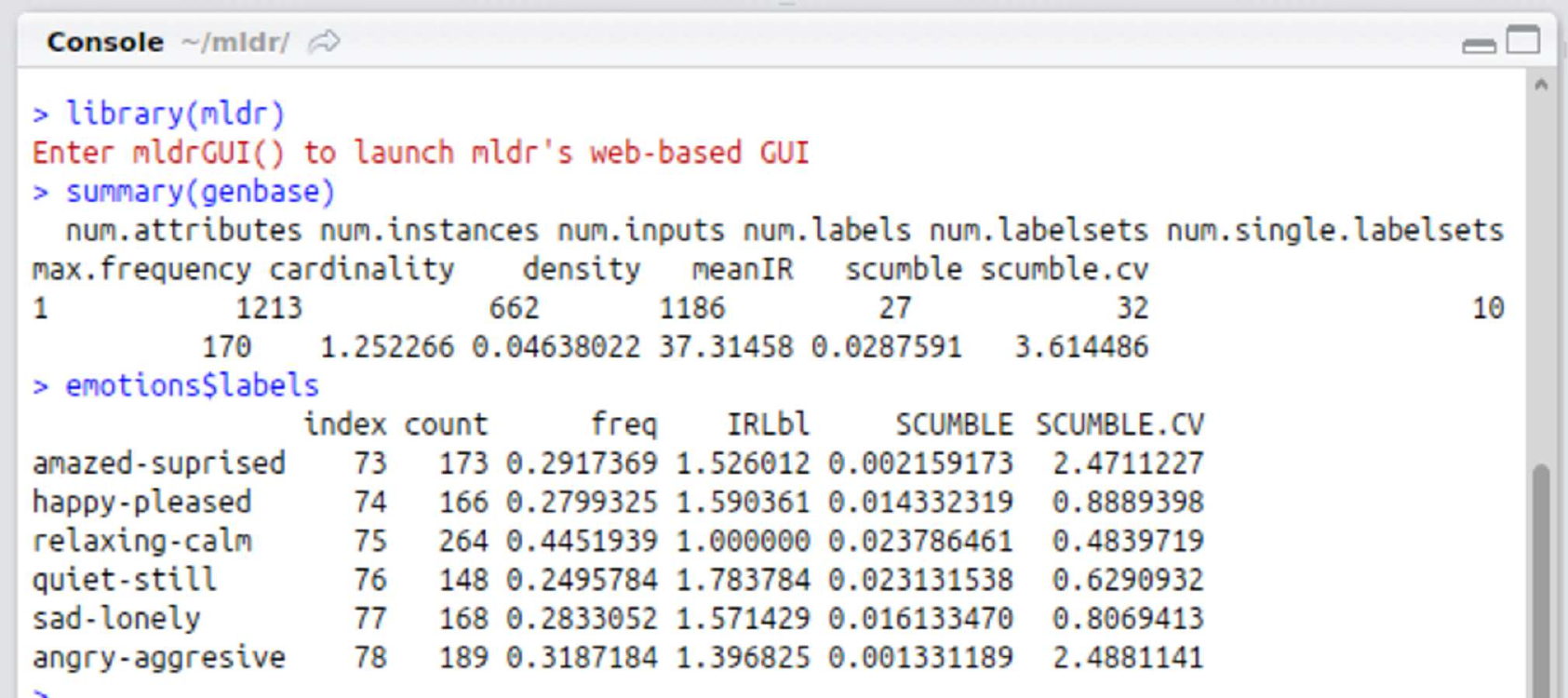}
\end{center}
\caption{Obtaining basic concurrence metrics using the mldr R package.} \label{mldrMetrics}
\end{figure*}

The group of metrics \textcolor {mycolor}{that} are general to the whole MLD can be retrieved with the \texttt{summary} function, as \textcolor {mycolor}{shown} in the upper part of Figure \ref{mldrMetrics}. In this example the measures belonging to the genbase MLD have been obtained. By querying the \texttt{labels} member of the object the information relative to each label is retrieved, including the \textit{SCUMBLELbl} and its corresponding coefficient of variation as shown in the bottom part of the same Figure \ref{mldrMetrics}.

Relying \textcolor {mycolor}{on} the measures obtained with the previous methods, essentially the \textit{IRLbl}, \textit{SCUMBLELbl} and name of each label, it is possible to infer which are the minority labels and which of those are more affected by the concurrence problem. However, it would not be easy to know what majority labels are interacting with each minority one. This information can be visually explored using the specific \texttt{plot} function provided by the mldr package, able to generate interaction plots similar to the ones shown in Figure \ref{genbase} and Figure \ref{yeast}.

\begin{figure*}[ht!]
\begin{center}
  \includegraphics[width=\textwidth]{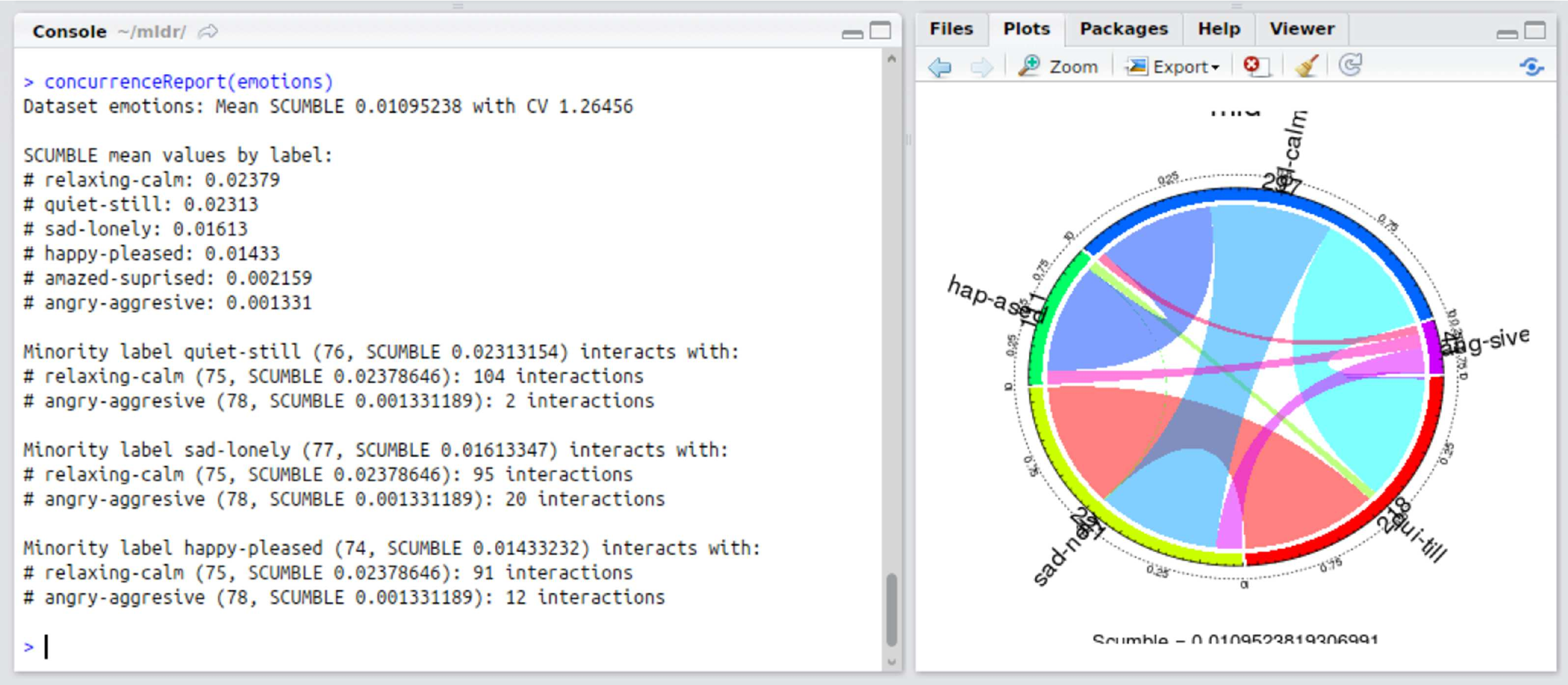}
\end{center}
\caption{The concurrence report provides information about label interactions, both textually and visually.} \label{mldrConcurrenceReport}
\end{figure*}

Another alternative would be calling the mldr's function \texttt{concurrenceReport}. It generates a full report stating what are the \textit{SCUMBLE} levels for the MLD and each of its labels, as well as a summary of label interactions and a plot of them. This report is sent to the console by default (see Figure \ref{mldrConcurrenceReport}), but it can also be saved as a PDF document by providing the \texttt{pdfOutput} parameter with the \texttt{TRUE} value. The report will include the minority labels most affected by the concurrence problem, sorted by their \textit{SCUMBLELbl} value. It will be, in fact, a list of difficult labels, along with the majority labels each one of them interacts with. 

\begin{figure*}[ht!]
\begin{center}
  \includegraphics[width=\textwidth]{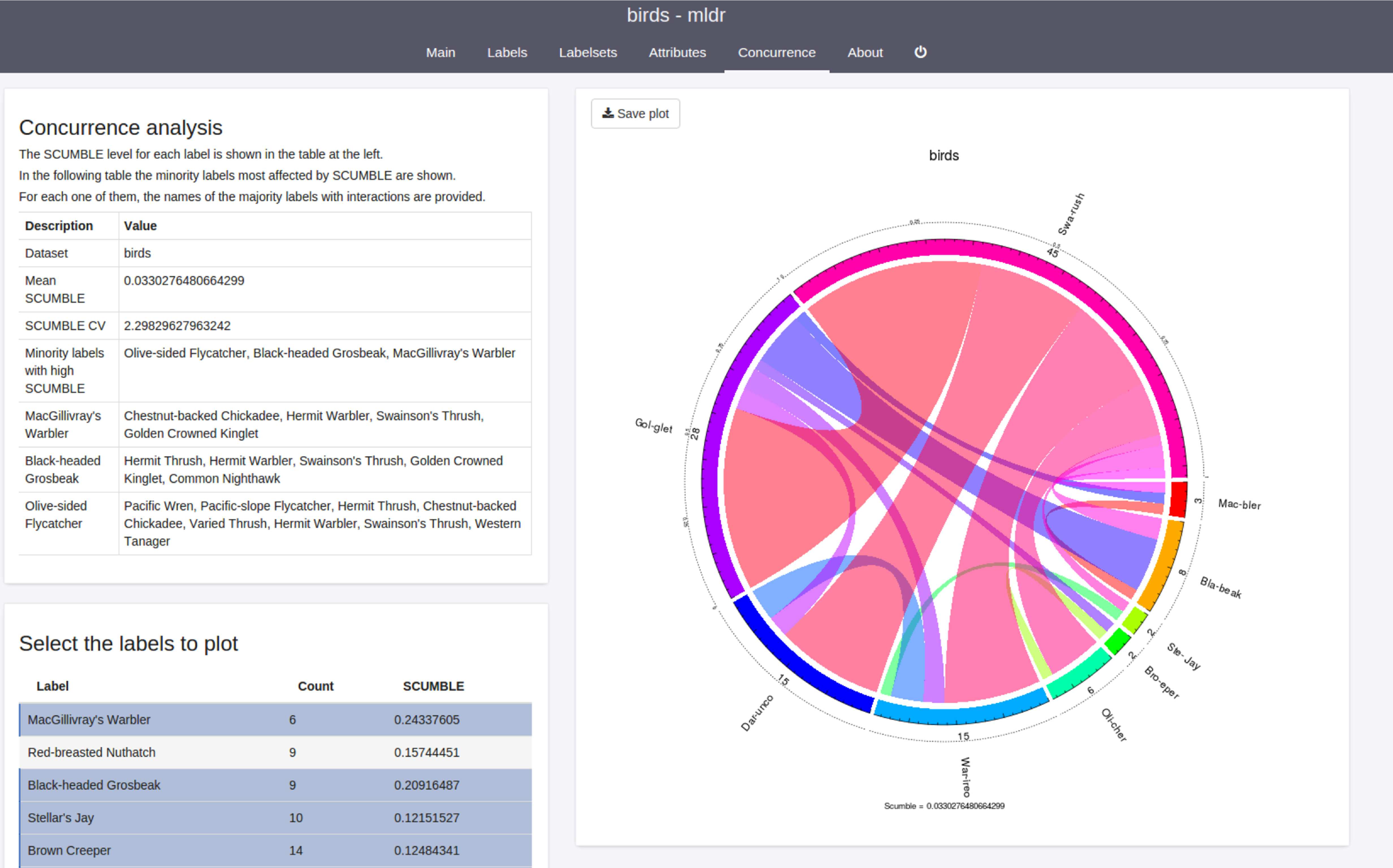}
\end{center}
\caption{The mldr GUI eases the process of obtaining customized concurrence information.} \label{mldrGUI}
\end{figure*}

In addition to the aforementioned functions, which are only a small sample of the set provided by the mldr package, a web GUI is also available. This can be launched from the R command line with the \texttt{mldrGUI}\footnote{An online version of the mldr's web interface, accessible from any browser without needing to install R or the mldr package, is available at \url{https://fdavidcl.shinyapps.io/mldr}. Although the bandwidth provided by shinyapps.io is limited, the application can be used to test the functionality described in this section.} function. It is structured into several pages, accessible by the tags located at the top. In the Concurrence page the same information provided by the \texttt{concurrenceReport} can be found, along with a customizable plot showing label interactions. This page is partially visible in Figure \ref{mldrGUI}. The list below the report allows the interactive selection of labels to be shown in the plot. The result can be saved to a file.

Overall, the exploratory tools implemented into the mldr package will provide all the information needed to analyze how the concurrence among imbalanced labels affects a certain MLD, as well as which of the labels could be considered difficult labels.

\subsection{The mldr package's REMEDIAL implementation}
\begin{figure*}[ht!]
\begin{center} \fbox{
  \includegraphics[width=\textwidth]{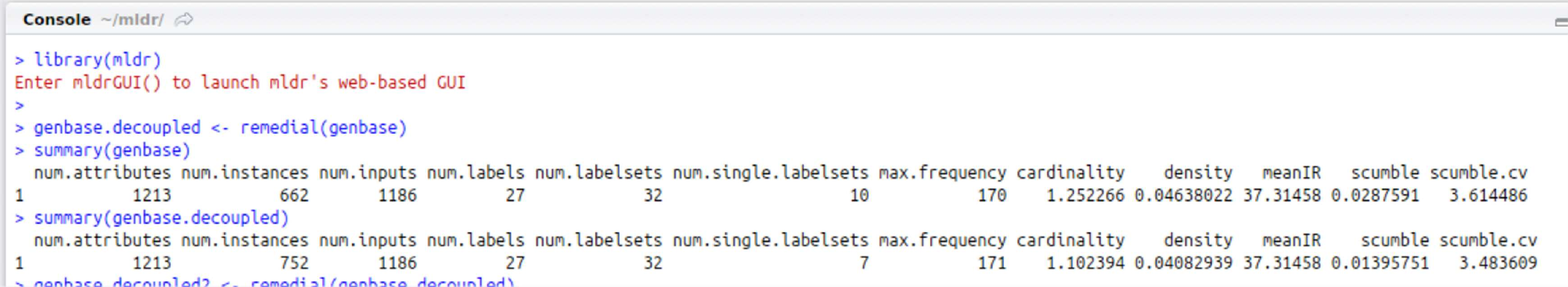}
  }
\end{center}
\caption{MLD basic traits before and after applying the REMEDIAL preprocessing algorithm.} \label{remedial}
\end{figure*}

\begin{figure*}[ht!]
\begin{center} \fbox{
  \includegraphics[width=\textwidth]{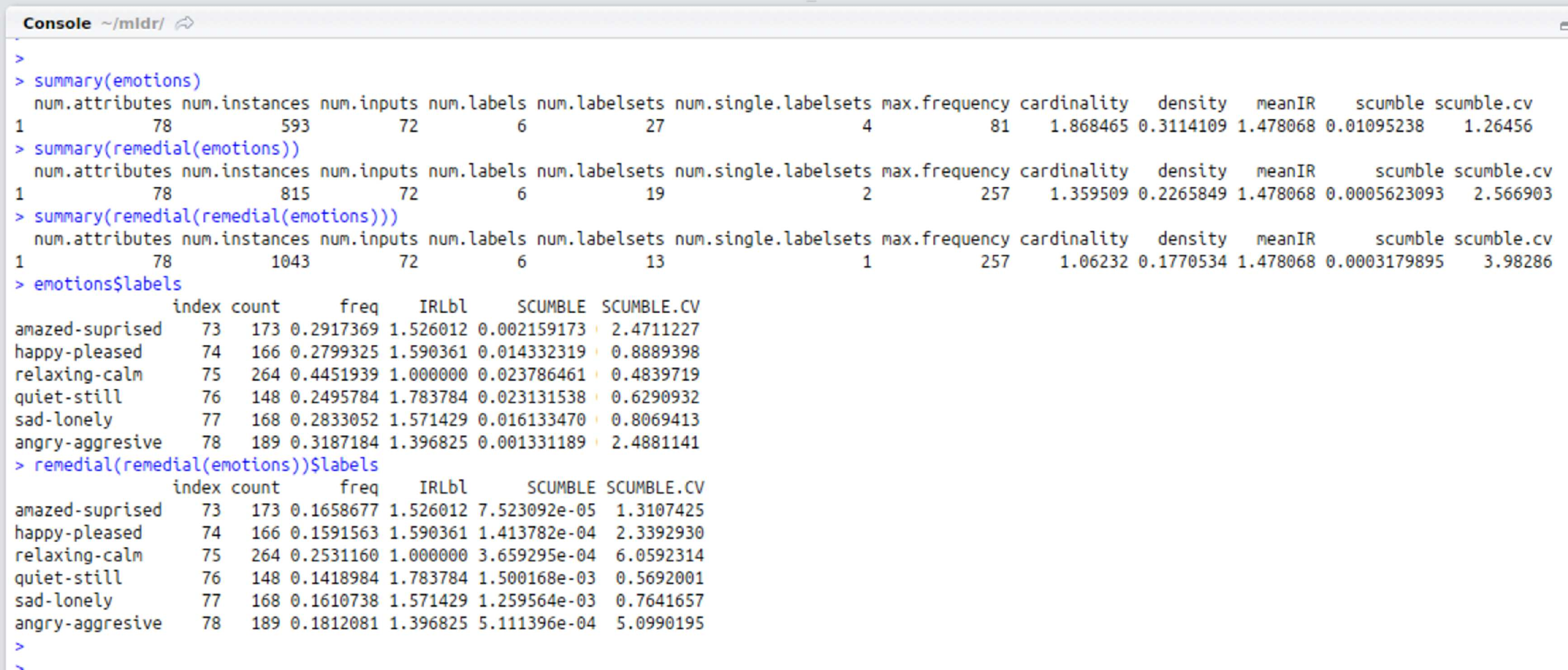}
  }
\end{center}
\caption{The algorithm can be applied several times to progressively reduce the concurrence problem.} \label{remedial2}
\end{figure*}

Along with the exploratory functionality previously described, the mldr package has been also extended by including a reference implementation of the algorithm REMEDIAL. The function containing this implementation is called \texttt{remedial}. To use it an mldr object has to be given as input, obtaining as output the preprocessed version of the same object. In Figure \ref{remedial} how to use this function is shown. The algorithm is applied to the genbase MLD, storing the result into the \texttt{genbase.decoupled} variable.

From the information provided by the \texttt{summary} function, corresponding to the MLD before and after the preprocessing, the following facts can be observed:
\begin{itemize}
    \item The number of instances grows, as REMEDIAL produces new data samples.
    \item Since the number of active labels in the MLD does not change, neither do the number of labels and the imbalance related metrics, such as \textit{MeanIR}.
    \item Because the same number of active labels are split into a larger number of instances, label cardinality and density decrease.
    \item In general, the decoupling of labels tend to produce simpler and more frequent labelsets. 
    \item The global \textit{SCUMBLE} and the \textit{SCUMBLELbl} are reduced.
\end{itemize}

As the algorithm REMEDIAL takes as reference the mean \textit{SCUMBLE} to determine which samples are going to be decoupled, and this measure is reduced as a result of applying REMEDIAL, it can be run several times over the same data to progressively reduce the concurrence problem. In Figure \ref{remedial2} the emotions MLD is used to show a simple example. The main metrics of the MLD and its labels are displayed after calling the \texttt{remedial} function once and twice. The differences are remarkable as can be seen.

\end{document}